\documentclass[sigconf]{acmart}
\acmSubmissionID{1039}

\usepackage{booktabs}

\setcitestyle{nosort,square} 



\copyrightyear{2024}
\acmYear{2024}
\setcopyright{rightsretained}
\acmConference[SIGGRAPH Conference Papers '24]{Special Interest Group on Computer Graphics and Interactive Techniques Conference Conference Papers '24}{July 27-August 1, 2024}{Denver, CO, USA}
\acmBooktitle{Special Interest Group on Computer Graphics and Interactive Techniques Conference Conference Papers '24 (SIGGRAPH Conference Papers '24), July 27-August 1, 2024, Denver, CO, USA}
\acmDOI{10.1145/3641519.3657508}
\acmISBN{979-8-4007-0525-0/24/07}

\usepackage{hyperref}
\usepackage{amsmath}
\usepackage{graphicx}
\usepackage{comment}
\usepackage{kbordermatrix}
\usepackage{multirow,bigdelim}

\usepackage{array}
\usepackage{wrapfig}

\usepackage{csvsimple}

\DeclareMathOperator*{\argmin}{arg\,min}

\usepackage{color}
\usepackage{soul}
\usepackage{booktabs}
\usepackage{tabularx}
\usepackage{rotating}
\usepackage{cleveref}
\usepackage{algpseudocode}
\usepackage{algorithm}

\definecolor{purple}{rgb}{0.99,0.2,0.72}
\definecolor{blue}{rgb}{0, 0.2, 0.8}
\definecolor{orange}{rgb}{0.6, 0.6, 0}
\definecolor{red}{rgb}{0.8, 0.2, 0.2}
\definecolor{magenta}{rgb}{0.5, 0.0, 1.0}
\definecolor{black}{rgb}{0.0, 0.0, 0.0}
\definecolor{cyan}{rgb}{0, 0.65, 0.65}
\definecolor{olive}{rgb}{0.2, 0.6, 0.5}
\usepackage{xcolor}

\newcommand{\X}{ \textbf{X} }
\newcommand{\A}{ \textbf{A} }
\newcommand{\MA} {\mathcal{A}}
\usepackage{dsfont}
\newcommand{\R}{ \mathds{R} }
\newcommand{\Ph}{\mathcal{P}}
\newcommand{\Pp}{\textbf{P}}
\newcommand{\Y}{\textbf{Y}}
\newcommand{\Q}{\textbf{Q}}
\newcommand{\F}{\textbf{F}}
\newcommand{\J}{\textbf{J}}

\newcommand{\fn}{\Psi}
\newcommand{\Loss}{\mathcal{L}}

\newif\ifdraft
\draftfalse

\ifdraft
\newcommand{\ssc}[1]{{\color{magenta}[\textbf{Sebastian:} \textit{#1}]}}
\newcommand{\yyc}[1]{{\color{olive}[\textbf{Yuting:} \textit{#1}]}}
\newcommand{\plc}[1]{{\color{blue}[\textbf{Peizhuo:} \textit{#1}]}}
\newcommand{\OSHc}[1]{{\color{purple}[\textbf{Olga:} \textit{#1}]}}

\else
\newcommand{\ssc}[1]{}
\newcommand{\yyc}[1]{}
\newcommand{\plc}[1]{}
\newcommand{\OSHc}[1]{}

\fi

\usepackage{pgfplots}
\pgfplotsset{compat=newest}
\usepgfplotslibrary{groupplots}
\usepgfplotslibrary{dateplot}

\usepackage[bottom]{footmisc}
\begin{document}

\title{WalkTheDog: Cross-Morphology Motion Alignment via Phase Manifolds}

\author{Peizhuo Li}
\orcid{0000-0001-9309-9967}
\affiliation{\institution{ETH Zurich} \country{Switzerland}}
\email{peizhuo.li@inf.ethz.ch}

\author{Sebastian Starke}
\orcid{0000-0002-4519-4326}
\affiliation{\institution{Meta Reality Labs} \country{United Kindom}}
\email{sstarke@meta.com}

\author{Yuting Ye}
\orcid{0000-0003-2643-7457}
\affiliation{\institution{Meta Reality Labs} \country{USA}}
\email{yuting.ye@meta.com}

\author{Olga Sorkine-Hornung}
\orcid{0000-0002-8089-3974}
\affiliation{\institution{ETH Zurich} \country{Switzerland}}
\email{sorkine@inf.ethz.ch}

\begin{abstract}
We present a new approach for understanding the periodicity structure and semantics of motion datasets, independently of the morphology and skeletal structure of characters. Unlike existing methods using an overly sparse high-dimensional latent, we propose a phase manifold consisting of multiple closed curves, each corresponding to a latent amplitude. With our proposed vector quantized periodic autoencoder, we learn a shared phase manifold for multiple characters, such as a human and a dog, without any supervision. This is achieved by exploiting the discrete structure and a shallow network as bottlenecks, such that semantically similar motions are clustered into the same curve of the manifold, and the motions within the same component are aligned temporally by the phase variable. In combination with an improved motion matching framework, we demonstrate the manifold's capability of timing and semantics alignment in several applications, including motion retrieval, transfer and stylization. Code and pre-trained models for this paper are available at \hyperlink{https://peizhuoli.github.io/walkthedog}{peizhuoli.github.io/walkthedog}.

\end{abstract}

\begin{teaserfigure}
    \centering
    \includegraphics[width=\linewidth]{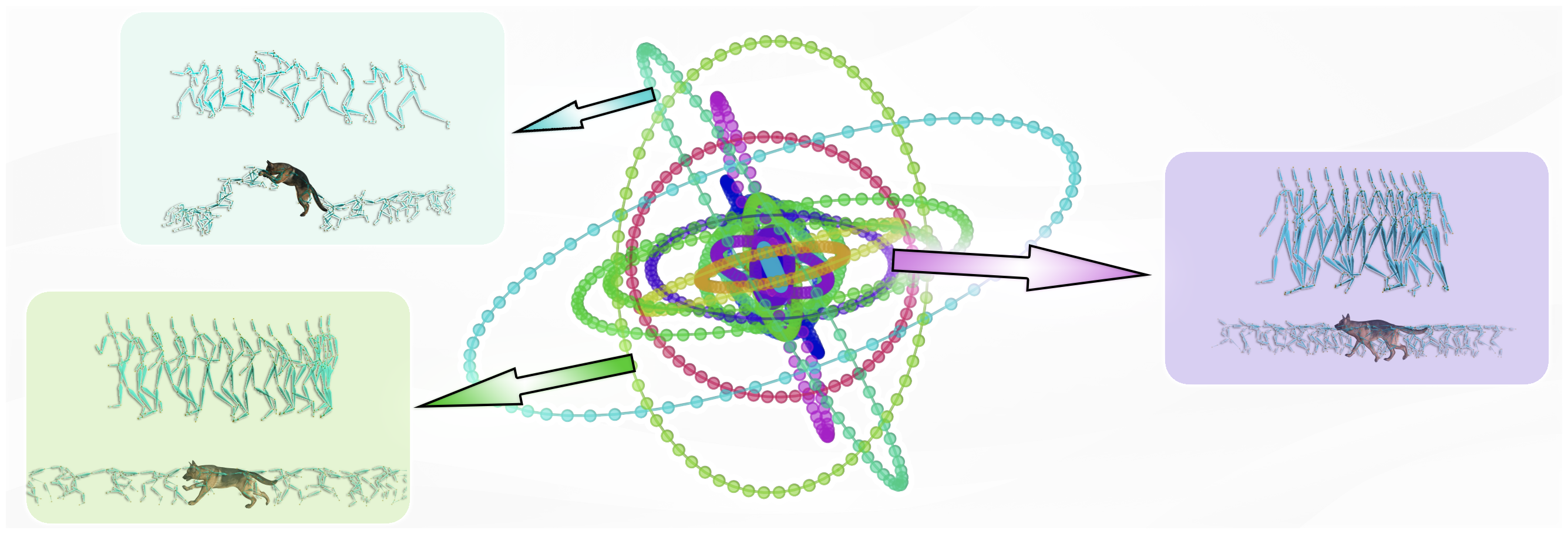}
    \caption{Our phase manifold $\Ph$ is learned from datasets with drastically different skeletal structures without any supervision. Each connected component in the manifold, visualized in a different color, is an ellipse embedded in high-dimensional space. Semantically similar motions from different characters are embedded into the same ellipse.}
    \label{fig:teaser}
    \label{fig:manifold}
\end{teaserfigure}

\maketitle

\section{Introduction}

What is in common between a dog's walk and a human's walk, or that of an ogre? Understanding the intrinsic structure and semantics of motion, regardless of the character's morphology and skeletal structure, lies at the heart of character animation research. In particular, motion retargeting and style transfer often rely on precise alignment of source and target motions in the form of paired data, posing severe limitations on their applicability. To make use of large heterogeneous datasets, common approaches organize motions in a discrete graph (motion graph \cite{kovar2002motion}) or a contiguous field (motion field \cite{motionfield}) with great success for many control and synthesis tasks. However, they fall short of handling different character designs and diverse content in a unified space. We argue that the main drawback of these methods is that their similarity metrics are based on \emph{extrinsic} pose features, which also encode features of the skeleton and motion semantics. In this work, we aim to learn an \emph{intrinsic} motion representation that is agnostic to the character morphology and can disentangle motion structure from semantics without any labels or other supervision signals.

An intrinsic property of motion is its periodic structure. Common locomotion such as walking and running can be effectively parameterized by a linear \emph{phase} variable for motion control problems \cite{holden2017phase, peng2018deepmimic}. 
To this end, we propose a latent representation that decomposes motions into a 1D phase and discrete amplitude vectors. This latent space forms a one-dimensional manifold that consists of multiple connected components, where each component is an ellipse corresponding to a discrete amplitude vector. We term it a \emph{disconnected 1D manifold}.
The possible choices of amplitudes are learned through vector quantization~\cite{van2017neural}, similar to a clustering process. The discrete amplitude vectors serve as a narrow bottleneck to regularize unsupervised learning of semantic motion clusters. The number of amplitude vectors reflects the semantic diversity of the motion dataset.

Formally, we propose a vector quantized periodic autoencoder (VQ-PAE) that embeds motions into a disconnected 1D manifold. The encoder projects a short input sequence into a 1D continuous phase variable and a latent code from a small codebook. The decoder reconstructs the input sequence using a simple two-layer convolution network with limited capacity to prevent memorization. The codebook and the autoencoder are jointly learned end-to-end. The small codebook size and the simple decoder enforce the semantic structure in the latent space. For example, idling and running will be far apart in the codebook because the decoder cannot reconstruct both from the same or similar input. On the other hand, jogging and running may have to share the same code or be close if the codebook size is small, as they are sufficiently similar when phase-aligned. When learning VQ-PAEs from multiple characters, such as a dog and a human, each character has their own VQ-PAE to handle their unique morphology and skeletal features, but they all share the same latent codebook. As a result, they are naturally clustered semantically as enforced by the codebook size, without any explicit supervision, but solely based on the intrinsic structure of the motion. Note that the VQ-PAE is not meant to be a generative model, given the intentional bottlenecks in the codebook and the decoder. We make use of the latent representation but discard the decoder after training.

We validate our design by learning VQ-PAEs from both a human dataset and a dog dataset with a shared codebook. Examining the average pose at each point on the manifold reveals that the learned embeddings are both timing- and semantics-aligned between the two characters. This highly structured and aligned phase manifold opens up new possibilities for motion data organization, retrieval, transfer and stylization. The phase manifold embedding can be flexibly integrated with existing motion synthesis pipelines. For example, given an unseen human motion, we can search the shared manifold for the nearest neighbor of dog motion with similar semantics and timing. We can further combine motion matching~\cite{motionmatching} with linear time warping supported by the 1D phase variable to transfer semantically similar motions between the human and the dog, without any paired data or pre-defined mapping among the skeletal structures. In addition, we demonstrate applications of motion characterization on the MOCHA dataset~\cite{jang2023mocha}.

Our key contributions are summarized as follows:
\begin{itemize}
    \item A novel phase manifold designed for both timing and semantics alignment. We also show that the manifold is compact, disentangled, and highly structured.
    \item A demonstration of using narrow bottlenecks and intrinsic structure of motions to achieve alignment among heterogeneous datasets, without any supervision, self-supervised losses, or skeletal structure correspondences.
    \item Applications with an improved motion matching framework on the phase manifold for motion retrieval, transfer and stylization.
\end{itemize}

\section{Related Work}
\begin{figure*}
    \centering
    \includegraphics[width=\linewidth]{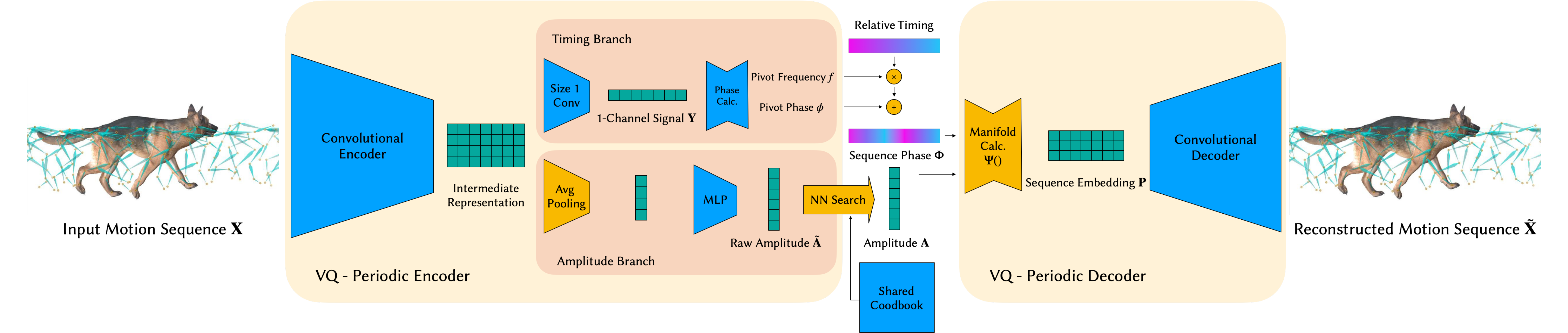}
    \caption{Architecture of VQ-PAE. Starting with a short motion sequence $\X \in \R^{J \times T}$, the encoder learns an intermediate representation using convolution. The representation is fed into the timing and the amplitude branch for predicting the phase $\phi$, the frequency $f$ and the amplitude $\A$ of the pivot frame (rendered with mesh). A vector quantization (i.e. nearest neighbor search) is used in the amplitude branch to ensure the structure of the phase manifold. Note the codebook $\mathcal{A}$ is shared among multiple VQ-PAEs. We calculate the embedding $\Pp$ of the sequence assuming the frequency and amplitude stay constant in the sequence. The predicted phase manifold sequence is then passed through a convolutional decoder to reconstruct the input motion. Components with learnable parameters are marked in blue.}
    \label{fig:arch}
\end{figure*}

In this section, we review the related work mainly on clustering and organizing motion capture datasets. We take a deeper look into the works related to \emph{phase} in terms of motion organization. Motion retargeting and style transfer are also related, in the sense of bridging different characters and distilling the core content of motions. We briefly review them at the end of this section. 

\paragraph{Organizing and clustering motion dataset}

Organizing a large-scale motion capture dataset is a difficult yet important task for 
applications. Graph-based methods~\cite{kovar2002motion, arikan2002interactive} find similar patterns of poses, cluster them into the same node, and use the edges to represent transition motions between nodes. This approach allows interactive control by mapping the user control to paths on the graph. \Citet{min2012motion} use key-frame-based segmentation to construct the graph structure and build probabilistic-based to increase the expressiveness and diversity of generated motion. At the same time, similar probabilistic models on graph structures are proposed. \citet{park2011finding} organize a motion capture dataset using context-free grammar learned from segments clustered with Partitioning Around Medoids (PAM) algorithm based on pose level similarity. \citet{aristidou2018deep} notice that semantic similarity may not be reflected by low-level representations such as poses and propose to learn a high-dimensional representation of motion motifs and motion signatures. 
Since the discrete structures lack expressiveness and responsiveness, \citet{motionfield} take another approach and learn a continuous field for motion. However, to generate motion, reinforcement learning is required to progress in the learned field. Motion matching~\cite{motionmatching} skips the organization of data and directly finds the best match of the current state and control signal in the dataset and replays the sequence. It is among the methods with the highest quality and is widely used in industry.
With the progress of tokenization~\cite{dhariwal2020jukebox, rombach2022high} with VQ-VAE~\cite{van2017neural}, it is becoming more and more popular for organizing human motion~\cite{geng2023human}, and demonstrated great success in multi-modal tasks~\cite{guo2022tm2t, siyao2022bailando, zhang2023t2m}. However, completely discretizing the latent space makes it difficult to capture the continuous nature of motion, and the learned latent space is usually less compact, making it difficult to construct a shared latent space for multiple characters.

\paragraph{Exploiting periodicity and phase}

Using phase and frequency domains to organize motion is closely related to our method. \citet{park2002line} propose to align motions by the key-frames such as foot-contact as key poses, and warping the motion with the guidance of key poses so motions at different speeds can be interpolated. It serves as an early inspiration for the introduction of phase and is part of the inspiration for our frequency-scaled motion matching in \Cref{sec:fs}. \citet{unuma1995fourier} demonstrate style transfer can be performed in the frequency domain. The introduction of \emph{phase} into neural networks demonstrated great success. It originally started with 1D phase~\cite{holden2017phase} coming from a semi-automated labeling process and quickly expanded into multiple dimension hand-crafted phase that is able to handle complex and non-periodic motions~\cite{starke2019neural}. \citet{starke2020local} attach a phase to each limb to deal with complex multiple contacts. DeepPhase~\cite{starke2022deepphase} proposes periodic autoencoders (PAEs), enabling learning on a continuous and expressive multi-dimensional phase manifold. It has been proven successful in applications like pose estimation~\cite{shi2023phasemp} and motion in-betweening~\cite{starke2023phasebetweener}. However, the learned phases and amplitudes are usually entangled, making it difficult to separate the timing and high-level semantics of motion. The sparsity of motion data leaves a large portion of the phase manifold invalid and can lead to implausible motions when used for synthesis, and it will be even more challenging to learn a shared phase manifold for multiple characters. We provide a comparison with DeepPhase of the disentanglement of phase manifolds in \Cref{sec:disentangle}.

\paragraph{Motion retargeting and style transfer}
\citet{gleicher1998retargetting} proposed one of the earliest method for motion retargeting, by directly optimizing on low-level motion representations. Other optimization-based methods~\cite{lee1999hierarchical,choi2000online, tak2005physically} are also proposed to improve the result. However, those methods mainly focus on transferring motions to a new skeleton, instead of building a common representation for different characters. This is only addressed with deep learning based methods~\cite{villegas2018neural, lim2019pmnet, aberman2020skeleton, li2023ace}, where a common latent space among different characters is learned. Although they may not need paired data, the same or homeomorphic skeletons are required such that the learning and auxiliary losses can be applied, while our method does not have this constraint. It is also demonstrated by \citet{kim2022humanconquad} that with paired examples, it is possible to retarget between bipeds and quadrupeds. In combination with the view of dynamic systems, \citet{kim2020learning} show that a common latent space for two similar dynamic systems for bipeds or pendulums can be learned with a pair of autoencoders. For style transfer, \citet{xia2015realtime} propose to use KNN search to build the style regression model. \citet{aberman2020unpaired} disentangles the style code and content code with a labeled dataset. \citet{jang2023mocha} make a further step to distinguish stylization and characterization, pushing the boundary of style transfer further. Our method can achieve a similar effect by treating each style as a separate dataset and using the alignment ability to transfer the content.

\section{Phase Manifold}

In this section, we introduce the design of our disconnected 1D phase manifold, which allows us to align motions with a single timing variable while creating a narrow bottleneck and forcing our framework to cluster semantically similar motions into the same connected component of the phase manifold. We then describe our vector quantized periodic autoencoder (VQ-PAE) to learn the embedding of motions from one dataset. Finally, we explain the approach for training multiple VQ-PAEs on different datasets into a common phase manifold.

\subsection{Disconnected 1D phase manifold}

We construct a phase manifold such that the timing is controlled by a 1D phase variable. Given an input motion sequence $\X \in \R^{J \times T}$, where $J$ and $T$ indicate the degree of freedom and the number of frames, respectively, we aim at mapping each frame $\X_i$ to a point $p = \fn(\A, \phi) \in \R^d$ on the phase manifold $\Ph$, parameterized by a 1D phase variable $\phi \in (-\frac12, \frac12]$ and a vector amplitude $\A \in \R^{2d}$. We choose the mapping $\fn$ to be
\begin{equation}
    \label{eq:phase_manifold}
    \fn(\A, \phi) = \A^0\sin(2\pi \phi) + \A^1 \cos(2\pi \phi) ,
\end{equation}
an ellipse embedded in $\R^d$, where $\A^0, \A^1 \in \R^d$ are the first and second half of $\A$, respectively. In contrast to $\phi$, which can take any value in $(-\frac12, \frac12]$, $\A$ can only be chosen from a finite codebook $\MA \subset \R^{2d}$ with size $K$. Thus, our phase manifold $\Ph$ can be formally defined as $\{\fn(\A, \phi)\ |\ \A \in \MA, \phi \in (-\frac12, \frac12]\}$. This construction gives us a latent space that is a collection of ellipses, as shown in \Cref{fig:manifold}, where we collect samples of $\Ph$ by uniformly sampling the phase $\phi$ on each ellipse $\Ph_i = \{\fn(\A_i, \phi)\ |\ \phi \in (-\frac12, \frac12]\}$ and use PCA to reduce dimension for visualization. In this manifold, a class of motions with similar semantics is embedded into the same ellipse. Note there is a one-to-one mapping between ellipses $\Ph_i$ and amplitudes $\A_i$. This allows us to flexibly scale the size of the bottleneck by changing the size of $\MA$. 
A properly chosen bottleneck size is the key to learning an expressive yet semantically aligned phase manifold.

\subsection{Vector quantized periodic autoencoder}

\citet{starke2022deepphase} introduce periodic autoencoder (PAE) for learning a continuous phase manifold. To learn a discrete amplitude space, we utilize the vector quantization technique to cluster the amplitude vectors into a learnable codebook $\MA$. The architecture of our vector quantized periodic autoencoder (VQ-PAE) is demonstrated in \Cref{fig:arch}. 

A desired mapping from motion to phase manifold should satisfy the following properties for an input motion sequence $\X \in \R^{J \times T}$ containing roughly a cyclic motion:

\begin{itemize}
    \item \emph{Phase linearity}: the phase $\phi$ should increase as linearly as possible over time.
    \item \emph{Amplitude constancy}: the amplitude $\A$ should be as constant as possible over time.
\end{itemize}
To achieve those two properties, we use a similar approach as PAE~\cite{starke2022deepphase} by using an encoder to predict the amplitude $\A$, the phase $\phi$ and the frequency $f$, which is the change rate of phase over time, at the center frame, \emph{i.e.} the \emph{pivot} frame, of a short input motion sequence $\X$. We then assume the two properties hold for the whole input sequence $\X$ and extrapolate the phase linearly with the predicted frequency to the whole sequence. We calculate the embeddings using \Cref{eq:phase_manifold} with extrapolated phases and amplitudes. A decoder is then used to reconstruct the input motion sequence from the predicted embedding. A decent reconstruction can only be achieved if the learned mapping is close to phase linear and amplitude constant.

\paragraph{Encoder} The encoder consists of a 2-layer 1D convolutional network mapping the input to an intermediate representation. The intermediate representation is then fed into two branches, namely the timing branch and the amplitude branch, each responsible for the prediction of phase, frequency and amplitude, respectively. We denote the relative timing of each frame in the sequence w.r.t. the pivot frame as $\mathcal{T} = \{t_i\}_{i=1}^T$, where $t_i = \left[ i - (T + 1)/2\right] \Delta_T$. Note we choose $T$ to be an odd number such that the pivot frame is unique, and $\Delta_T$ is the frame time of the dataset.

\begin{figure}
    \centering
    \includegraphics[width=\linewidth]{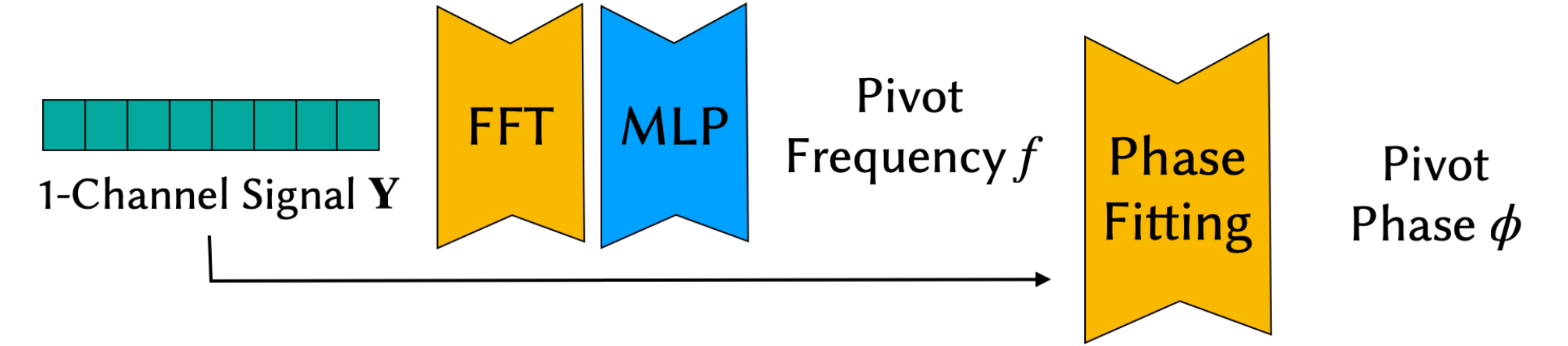}
    \caption{Details of phase calculation module.}
    \label{fig:phase_calculation}
\end{figure}

\paragraph{Timing branch} The timing branch starts with a 1D convolution with kernel size 1, mapping the multi-dimensional intermediate representation to a 1-channel temporal signal. A phase calculation module is used on the temporal signal to predict the phase $\phi$ and frequency $f$. The detailed architecture of the phase calculation module is shown in \Cref{fig:phase_calculation}. PAE~\cite{starke2022deepphase} uses the power of each frequency bin calculated by fast Fourier transform (FFT) as weights to calculate the average frequency. However, it produces unstable frequencies as the input phase shifts even when it is a sinusoidal signal with a non-integer frequency. We find that using a small multi-layer perceptron (MLP) on the powers produces more robust frequency prediction. We use the equations presented by~\citet{IanWebPost} to calculate the phase $\phi$, which helps with the fact that $\phi$ is not a continuous parameterization of the phase manifold. Please refer to the supplementary material for more details.

\paragraph{Amplitude branch} As the amplitude should be nearly constant over time, we first apply an average pooling on the temporal axis on the intermediate representation. An MLP is followed to get a raw prediction of amplitude $\tilde\A$. Since the possible choices of amplitude are finite, we use a vector quantization layer to find the nearest neighbor $\A = \argmin_{\A_i \in \MA} \| \tilde\A - \A_i \|_2$.

\paragraph{Decoder} With phase linearity and amplitude constancy assumptions, the phase variable of the input motion can be calculated by $\Phi = \phi + f \cdot \mathcal{T}$ with the relative timing $\mathcal{T}$. The embedding of the input motion sequence can then be calculated by $\Pp = \fn(\A, \Phi)$. The decoder is a 2-layer 1D convolutional network that maps the embedding back to the original motion space. 

\paragraph{Loss function} We use the following loss function to train our VQ-PAE:
\begin{equation}
    \label{eq:loss}
    \begin{aligned}
        \mathcal{L}_{\text{rec}} &= \| \X - \tilde\X \|_2, \\
        \mathcal{L}_{\text{vq}} &= \|\text{sg}( \tilde\A) - \A \|_2 + \|\tilde\A - \text{sg}(\A) \|_2,
    \end{aligned}
\end{equation}
where $\tilde\X$ is the reconstructed motion sequence, $\text{sg}(\cdot)$ is the stop gradient operator. The first loss is the reconstruction loss of the VQ-PAE the second loss is the vector quantization loss~\cite{van2017neural}. The total loss is 
\begin{equation}
    \label{eq:total_loss}
    \mathcal{L} = \mathcal{L}_{\text{rec}} + \lambda_{\text{vq}} \mathcal{L}_{\text{vq}},
\end{equation}
and $\lambda_{\text{vq}}$ is a hyperparameter. For a detailed network architecture and hyperparameter settings, please refer to the supplementary material.

\subsection{Learning a common phase manifold among VQ-PAEs}

\begin{figure}
    \centering
    \includegraphics[width=\linewidth]{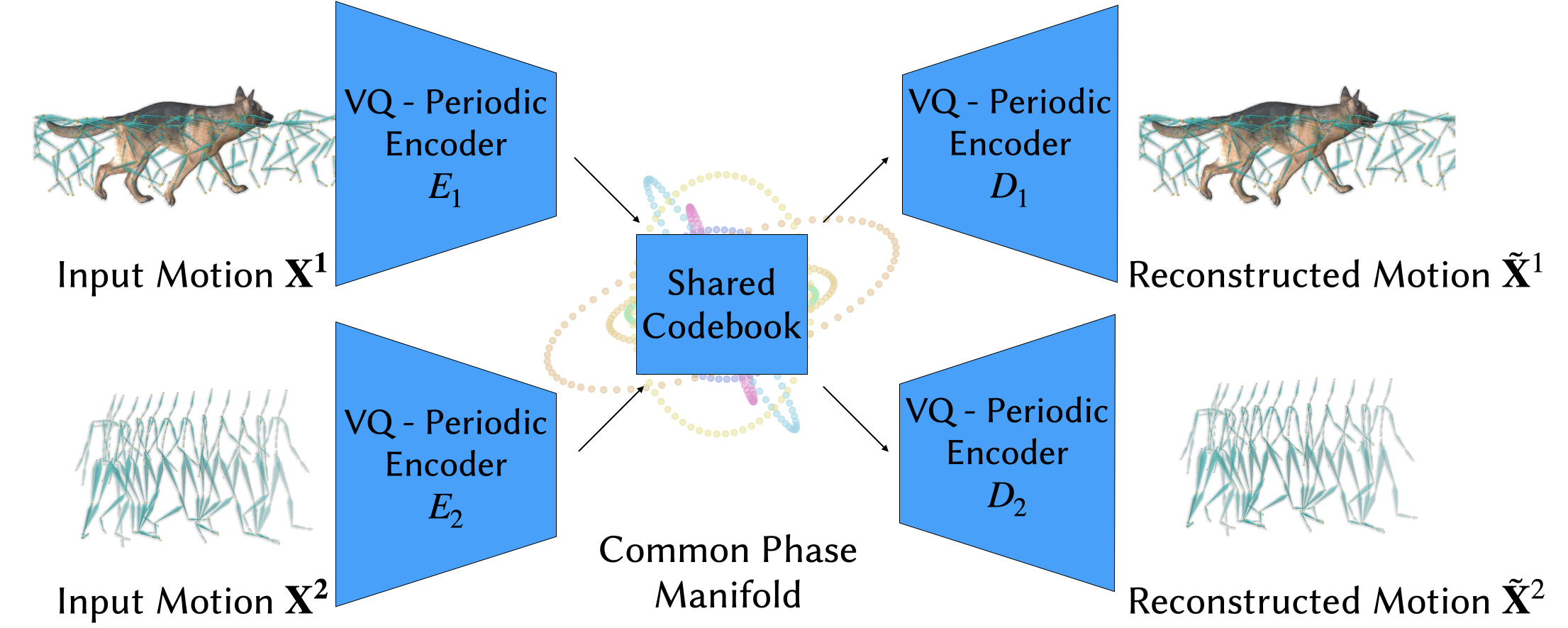}
    \caption{Overview of training multiple VQ-PAEs on heterogeneous datasets. A common phase manifold is guaranteed by using a shared codebook $\MA$.}
    \label{fig:joint_training}
\end{figure}

To align motions among different datasets, a common phase manifold can be learned with a shared codebook $\MA$ and no additional supervision as shown in \Cref{fig:joint_training}. Without loss of generality, we illustrate the training process of two VQ-PAEs on two datasets $\mathcal{D}_1$ and $\mathcal{D}_2$ with different skeletal structures in this section. The training process can be easily extended to more datasets.

The loss for training two VQ-PAEs can be written as
\begin{equation}
    \label{eq:joint_training}
    \Loss = \Loss_{\text{rec1}} + \Loss_{\text{rec2}} + \lambda_{\text{vq}} \Loss_{\text{vq}},
\end{equation}
where $\Loss_{\text{rec1}}$ and $\Loss_{\text{rec2}}$ are the reconstruction losses of the two VQ-PAEs, and $\Loss_{\text{vq}}$ is the vector quantization loss of the shared codebook $\MA$. During training, we optimize two VQ-PAEs at the same time. Note that we do not need any skeletal topology correspondences due to the use of simple 1D convolution and MLPs.

However, directly optimizing \Cref{eq:joint_training} can lead to situations where part of the entries in $\MA$ are only used by one VQ-PAE, causing disparity in the embeddings of $\mathcal{D}_1$ and $\mathcal{D}_2$. This is also a common problem for regular VQ-VAEs that many entries in the codebook are not used. \citet{zheng2023online} propose a simple yet effective reinitialization technique to solve this problem for training with one VQ-VAE. We adapt their method to the training of multiple VQ-PAEs.

\paragraph{Reinitialization of $\MA$} At the beginning of training, $\MA$ is initialized with uniform ditribution $\mathcal{U}[-1/K, 1/K]$ and $K = |\MA|$. 
For simplicity, we discuss the reinitialization of $\MA$ for one VQ-PAE.
At each training iteration step, the decayed running average usage $N_{i}^{(t)}$ at the $t$-th iteration of each entry $\A_i$ in $\MA$ by the  VQ-PAE is updated by 
\begin{equation}
    \label{eq:running_average}
    \textstyle
    N_{i}^{(t)} = \gamma N_{i}^{(t-1)} + (1 - \gamma) \frac{n_{i}^{(t)}}{N},
\end{equation}
where $n_{i}^{(t)}$ is the number of times $\A_i$ is used by the VQ-PAE at the $t$-th iteration, $N$ is the number of amplitudes produced by the encoder being quantized at each iteration and $\gamma$ is the decay rate. Intuitively, entries with low usage are more likely to be reinitialized. We choose to reinitialize the less frequently used entries to a randomly chosen amplitude produced by the encoder. Formally, the reinitialization target $Z_{i}$ of entry $\A_i$ is sampled such that closer outputs are preferred to maximize the utilization of the codebook by
\begin{equation}
    \label{eq:reinit_target}
    \mathbb{P}(Z_{i} = \tilde A_{k}) \propto \exp(-\|\A_i - \tilde\A_k\|_2),
\end{equation}
where $\{\tilde\A_k\}$ are the raw amplitudes predicted by the encoder in this iteration. At an update step, every entry in the codebook is linearly interpolated to the reinitialization target with a weight $\alpha_{i}$ by
\begin{align}
\textstyle
    \alpha_{i} &= \exp\left(-N_{i} \frac{10}{1 - \gamma} - \epsilon\right), \label{eq:reinit_alpha}\\
    \A_i &= (1 - \alpha_{i})\A_i + \alpha_{i} Z_{i}, \label{eq:reinit_val}
\end{align}
where $\epsilon$ is a small constant acting as a regularizer. We set $\alpha_i$ such that less frequently used $\A_i$ is interpolated more towards a randomly picked output of the encoder. Note the temporal superscript $(t)$ is omitted for simplicity. Since the codebook is shared among multiple VQ-PAEs, the reinitialization of $\MA$ is performed as the average update of all VQ-PAEs produced by \Cref{eq:reinit_val}. An entry will converge to a stable value only if it is frequently used by all VQ-PAEs. For more details and reasoning of the setting of $\epsilon$ and $\gamma$, we refer the readers to the work of \citet{zheng2023online}. $\MA$ is reinitialized at every training iteration before the gradient descent step.

Existing methods for learning a common latent space for motions with different skeletons~\cite{villegas2018neural, aberman2020skeleton} usually require at least partially specified skeletal topology correspondences and additional implicit supervision such as cycle consistency~\cite{zhu2017unpaired} and adversarial training~\cite{goodfellow2020generative}. In contrast, our method achieves a common phase manifold with only a shared codebook $\MA$ and no additional supervision, while semantics and timing alignment are naturally provided.
 Relying on the intrinsic periodicity of motions, this phase manifold can be used to model different character topologies including biped and quadruped without extra class-specific designs.

\section{Frequency-Scaled Motion Matching}
\label{sec:fs}

After the training of our VQ-PAEs, we can obtain the corresponding manifold embedding $p_i \in \Ph$ for every frame $i$ in the dataset, by using the encoder to encode a 1-second motion sequence centered at frame $i$. Although relying on a single point on the manifold to represent a pose can be ambiguous, since the manifold is designed to be compact, a sequence of manifold points contains rich information to retrieve a motion sequence from the database. In fact, within a single cycle, the possible progress of phase, characterized by all possible mappings from time to phase $g \colon [0, 1] \to (-\frac12, \frac12]$, is very expressive. To exploit the expressiveness in a sequence, we demonstrate that it is possible to use motion matching~\cite{motionmatching} on the phase manifold and improve it with the explicit phase variable.

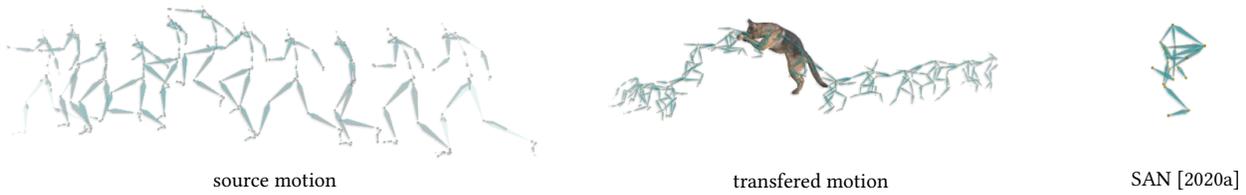
\begin{figure*}
    \centering

    \begin{tikzpicture}
        \newcommand{\pll}{-0.1}
        \node[anchor=south west,inner sep=0] (image) at (0,0) {\includegraphics[width=\linewidth]{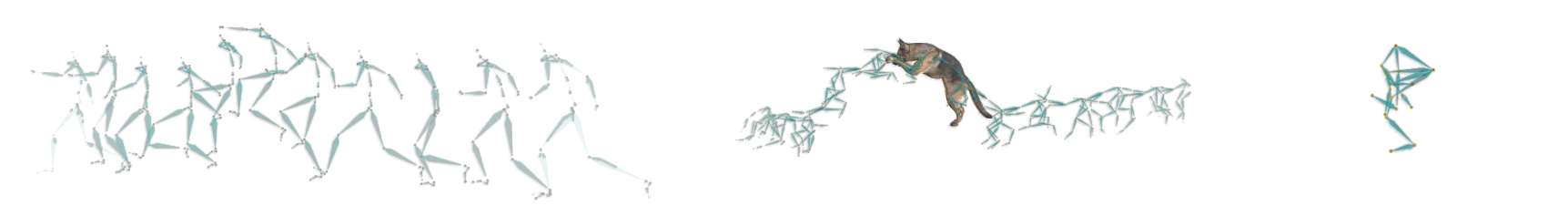}};
        \begin{scope}[x={(image.south east)},y={(image.north west)}]
            \small
            \node at (0.22, \pll) {source motion};
            \node at (0.62, \pll) {transfered motion};
            \node at (0.9, \pll) {SAN~\shortcite{aberman2020skeleton}};
        \end{scope}
    \end{tikzpicture}
    
    \caption{Motion transfer. Our framework can transfer motions between different characters preserving the semantics. However, SAN~\shortcite{aberman2020skeleton} produces implausible results because of unstable adversarial training.}
    \label{fig:transfer}
\end{figure*}

Given the phase embedding sequence $\Pp$ of an input motion sequence, we use motion matching to retrieve a motion sequence from the database, with phase embedding as the control signal in the classical motion matching algorithm. For more details of the implementation, please refer to the supplementary material.
We also compare the result performance of motion transfer on the dog-human setup with skeleton-aware networks (SAN)~\cite{aberman2020skeleton}, the state-of-the-art for skeletal motion retargeting between different skeletal structures. 
SAN heavily requires end-effector velocity consistency between the source and target characters and struggles to transfer motion with a large difference in skeletal structure as shown in \Cref{fig:transfer}.

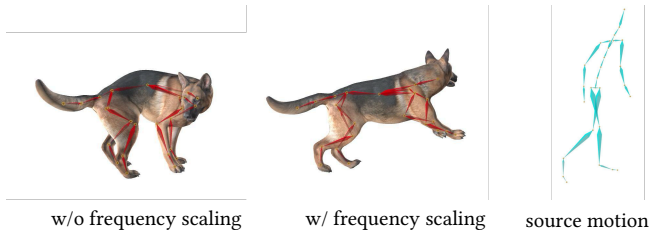
\begin{figure}
    \centering

    \begin{tikzpicture}
        \newcommand{\pll}{-0.1}
        \node[anchor=south west,inner sep=0] (image) at (0,0) {\includegraphics[width=\linewidth]{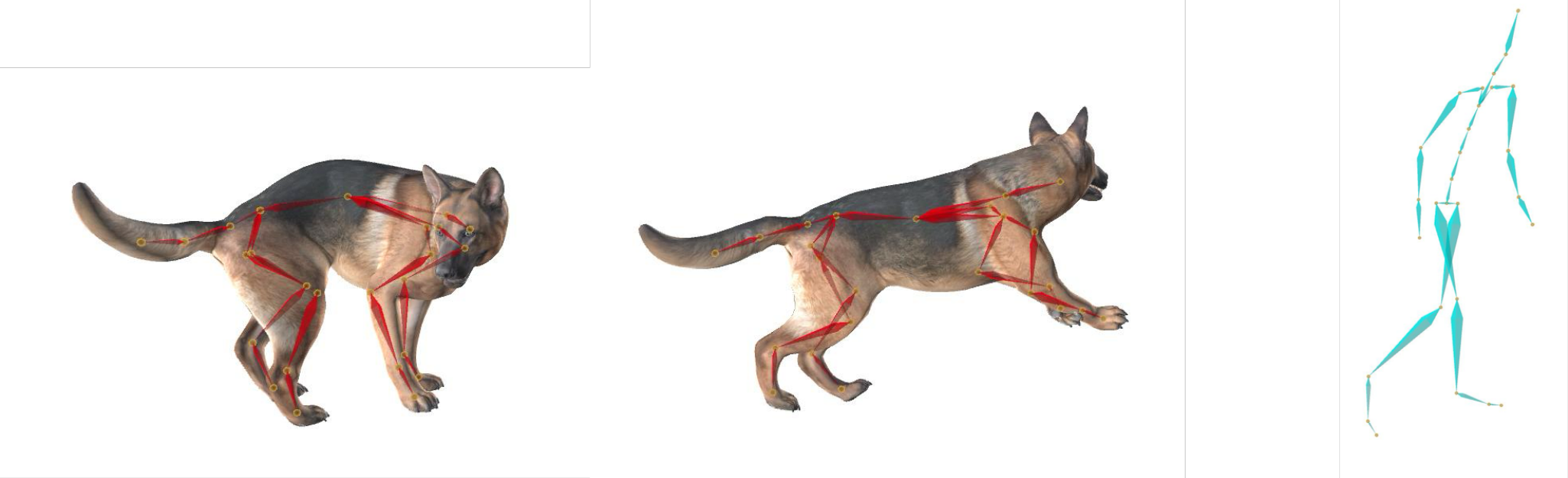}};
        \begin{scope}[x={(image.south east)},y={(image.north west)}]
            \small
            \node at (0.22, \pll) {w/o frequency scaling};
            \node at (0.61, \pll) {w/ frequency scaling};
            \node at (0.91, \pll) {source motion};
        \end{scope}
    \end{tikzpicture}
    
    \caption{The running motions in Dog and Human-Loco dataset are of different frequencies. With frequency scaling, the motion with correct semantics is matched.}
    \label{fig:semantics_misalignment}
\end{figure}

A common problem in motion matching is there is a trade-off between responsiveness and smoothness. This can be mitigated by using variable replay lengths $T_0$ depending on the control signal. However, this requires a lot of manual tuning and is not robust to different inputs. In addition to this problem, directly applying vanilla motion matching for motion transfer is not ideal, as there might not be a motion clip in the database sharing the same semantics and frequency as the input motion, causing timing or semantics misalignment.

\begin{algorithm}
    \caption{Frequency-scaled motion matching}
    \label{alg:fs}
    \begin{algorithmic}
        \State $i \gets 1$
        \State $\J_{\text{start}} \gets$ initial pose descriptor
        \While {$i < T$} 
            \State $k = \argmin_{k} \text{c}(i, k)$
            \State \emph{Output} $\Y_{k:k+t(k)}$ linearly interpolated to length $t(i)$
            \State $i \gets i + t(i)$
            \State $\J_{\text{start}} \gets \J_{k + t(k)}$
        \EndWhile
    \end{algorithmic}
\end{algorithm}

With the help of our phase manifold, we can solve both problems by performing matching on a fixed number of cycles instead of a fixed number of frames. We demonstrate the details with 1 cycle and this can be easily extended to arbitrary cycles.
Given a motion sequence $\X$ and its corresponding frequencies $\F = \{f_i\}$ predicted by the VQ-PAE, for each starting frame $i$, we define its period $t(i)$ as the first frame $j$ such that $\sum_{k=i}^{i+j} f_k \Delta_T \geq 1$, thus $\X_{i:i + t(i)}$ roughly corresponds to one cycle of motion. During motion matching, instead of using a fixed number of frames $T_0$, we query every period of the input manifold, while the query is conducted on sequence with 1-period length in the database, as shown in \Cref{alg:fs} and \Cref{eq:cost_fs}. We denote the phase sequences of the database with $\Q$, the pose descriptor used to measure the similarity between frames with $\J$ and the pose with $\Y$.
As a result, when more agile motions, \emph{i.e.} motions with higher frequency and lower period, are involved, the matching steps will be carried out more frequently and thus the motion will be more responsive. On the other hand, by allowing interpolating the output motion to the same frequency as the input, we achieve a more accurate timing and semantics alignment, as shown in \Cref{fig:semantics_misalignment}. 
The transition cost function $\text{c}(i, k)$ is defined as:
\begin{equation}
    \label{eq:cost_fs}
    \text{c}(i, k) = d(\Pp_{i:i+t(i)}, \Q_{k:k+t(k)}) + \lambda_1 \|\J_{\text{start}} - \J_k\|^2_2 + \lambda_2 \|t(i) - t(k)\|^2,
\end{equation}
where $d(\Pp, \Q)$ can be calculated by linearly interpolating their phases to the same length, chosen to be $1/\Delta_T$, and calculating the squared distance between them. The third term is introduced because we favor the motion with similar frequency and discourage large temporal interpolation. Note that $t(i)$ in the database and the fixed length interpolation of $\Q_{i:i+t(i)}$ can be precomputed, so the commonly used acceleration techniques for motion matching can still be applied to speed up the search.

\section{Applications and Evaluations}
\label{sec:experiments}

We evaluate our disconnected 1D phase manifold in terms of timing alignment and semantic alignment on several datasets. We show that our method can be used for improving motion matching with the predicted 1D phase. With our improved motion matching, we show that it is possible to achieve motion transfer and motion stylization by performing motion matching on the phase manifold. 

\subsection{Datasets}

We use three datasets in our experiment. The Dog dataset~\cite{zhang2018mode} and Human-Locomotion dataset~\cite{starke2019neural} contain mostly locomotion including walking, running, jumping and idling. The MOCHA dataset~\cite{jang2023mocha} is a recently proposed highly stylized and characterized motion dataset. It contains a wide range of motions on different characters, including clown, ogre, princess, robot and zombie. For a detailed demonstration of the dataset, we refer the readers to~\citet{jang2023mocha}. 
In the following sections, we train our VQ-PAEs with two combinations of datasets: Dog and Human-Locomotion and MOCHA-Clown and MOCHA-Ogre. We refer to the former as \emph{human-dog} setting and the latter as \emph{stylized} setting. 
In addition, we show that it is possible to learn a shared latent space for multiple datasets with different characters, such as Dog, Human-Locomotion, and MOCHA by extending \Cref{eq:joint_training} with additional reconstruction losses and training multiple VQ-PAEs together. Please refer to 3:10 in the accompanying video for a demonstration.

\subsection{Motion alignment}
\label{sec:motion_alignment}

We examine the average pose at each point of the manifold to verify its alignment effect. Since our 1D phase manifold is a compact embedding of motions, the mapping from $p_i$ to pose space is naturally a one-to-many mapping. However, it is not trivial to obtain the average on a continuous space. We propose to train a small MLP for each dataset that minimizes the following loss:
\begin{equation}
    \Loss_{\text{pose}} = \mathbb{E}_{(p_i, \Y_i) \sim \mathcal{D}_k}\|\Y_i - M_k(p_i)\|_2,
\end{equation}
where $M_k$ is the MLP for dataset $\mathcal{D}_k$ that maps a point in $\Ph$ to pose space, and $(p_i, \Y_i)$ are pairs of manifold embedding and the corresponding pose in the dataset $\mathcal{D}$. 

We uniformly sample phase variables with different amplitudes to get the embeddings and use the learned MLP to predict the corresponding poses. The results are shown in \Cref{fig:manifold}. It can be seen that even for drastically different characters like a dog and a human, where neither the semantic nor the timing alignment is well defined, the average poses from different datasets at the same manifold point provide a reasonable alignment on the semantic level. This is only possible if semantically similar motions are mapped into the same amplitude and poses with similar timing are mapped into the same phase, otherwise, the average poses would be noisy and meaningless. For more results, please refer to the accompanying video.

\subsection{Disentangling phase and amplitude}
\label{sec:disentangle}

\begin{figure*}
    \centering
    \includegraphics[width=\linewidth]{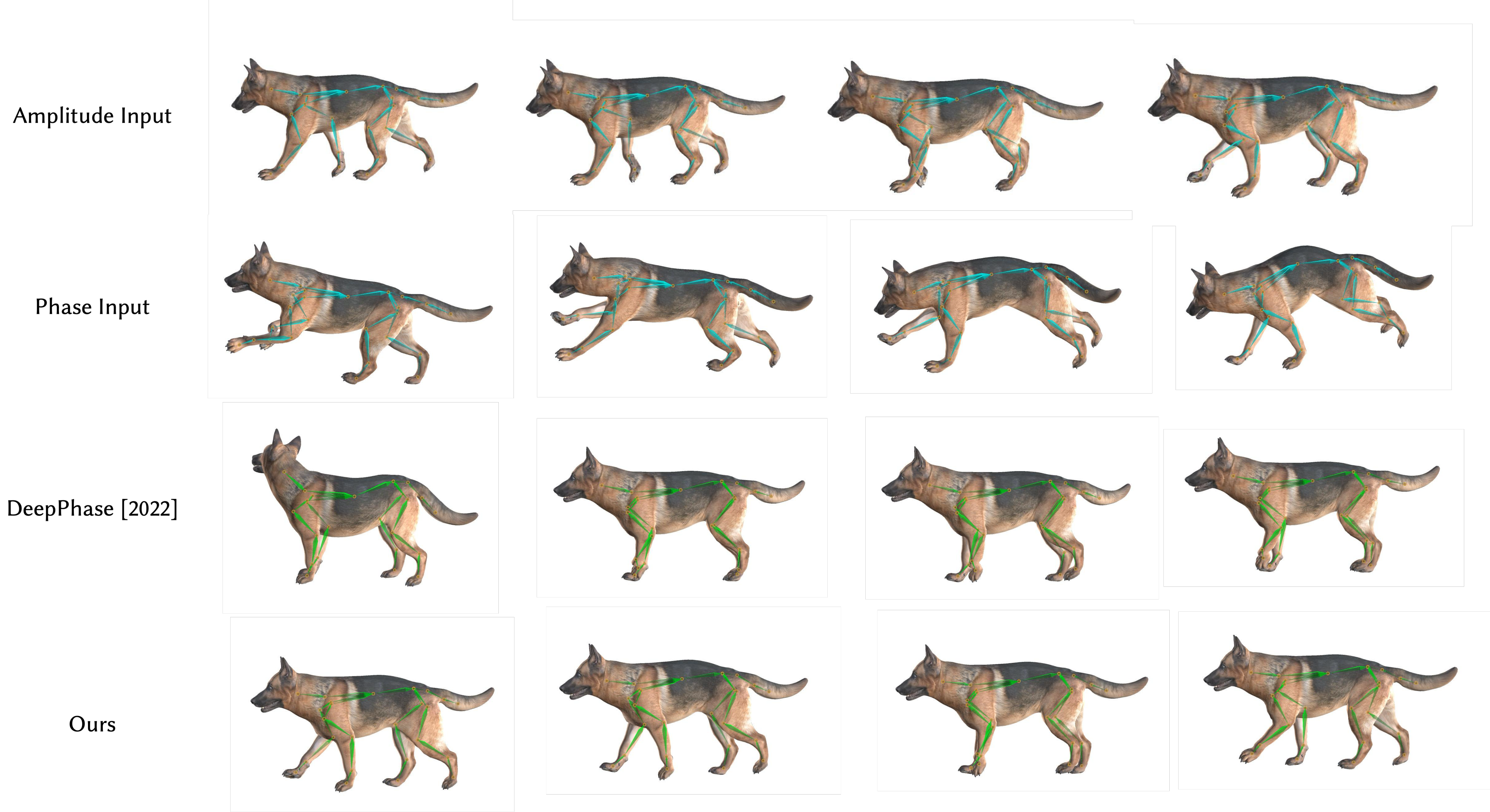}
    \caption{Phase and amplitude disentanglement. Our method generates motion combining the semantics from the amplitude input and the timing from the phase input, while DeepPhase~\shortcite{starke2022deepphase} generates implausible motions due to the entangled phase manifold.}
    \label{fig:disentangle}
\end{figure*}

In both phase manifolds designed by us and by DeepPhase~\cite{starke2022deepphase}, the phase represents timing and the amplitude represents motion content. We examine the phase-amplitude entanglement by training the same MLP mapping from the phase manifold to pose space as in \Cref{sec:motion_alignment}. By taking the amplitude from one motion sequence or a static pose and the phase from another motion sequence, we predict the corresponding pose using the trained MLP. It can be seen in \Cref{fig:disentangle} that our method can learn a disentangled phase manifold, but the manifold from DeepPhase fails due to the entanglement and non-compactness in using a multi-dimensional phase.

\subsection{Motion retrieval}

\begin{figure}
    \centering
    \includegraphics[width=\linewidth]{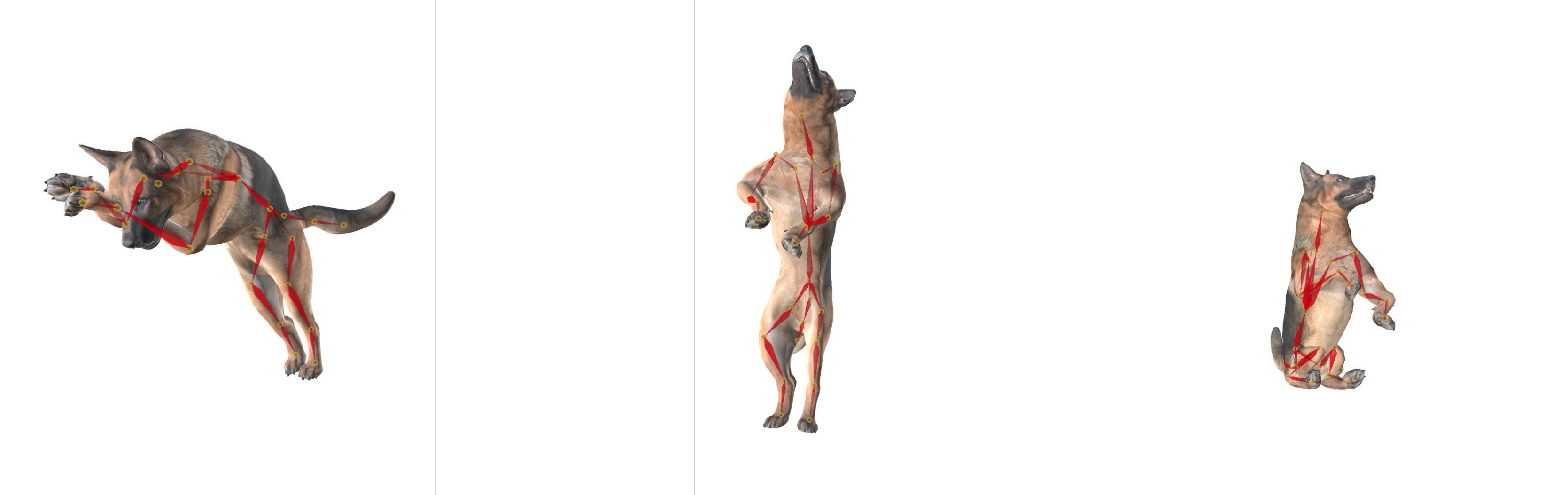}
    \caption{Motion retrieval. We retrieve motions at different frequencies in the same connected component containing motions of a dog moving up and down. From left to right the frequency decreases, corresponding to fast jumping, jumping up and sitting back, and slowly standing up and sitting back. Please refer to 1:17 in the accompanying video for a more comprehensive result.}
    \label{fig:retrieve}
\end{figure}

We show a simple example that by varying the frequency $f$, we can retrieve semantically similar motion at different frequencies by searching the nearest neighbor in the phase embeddings of the dataset, as shown in \Cref{fig:retrieve}. Formally speaking, given an amplitude $\A \in \MA$ and a frequency $f$, we generate a uniformly distributed phase sequence $\Phi_f = \{\phi_i\}_{i=1}^N$ with $\phi_i = i f \Delta_T$, and $N = 1/(f\Delta_T)$ such that $\Phi$ covers exact one cycle with frequency $f$. We then retrieve the desired motion with nearest neighbor search by comparing the constructed embedding sequence $\fn(\A, \Phi_f)$ and the embedding sequences of the motions with length $N$ from the dataset. Please refer to the accompanying video for a detailed result.

\subsection{Motion stylization and characterization}

\begin{figure*}
    \centering

    \begin{tikzpicture}
        \newcommand{\pll}{-0.1}
        \node[anchor=south west,inner sep=0] (image) at (0,0) {\includegraphics[width=\linewidth]{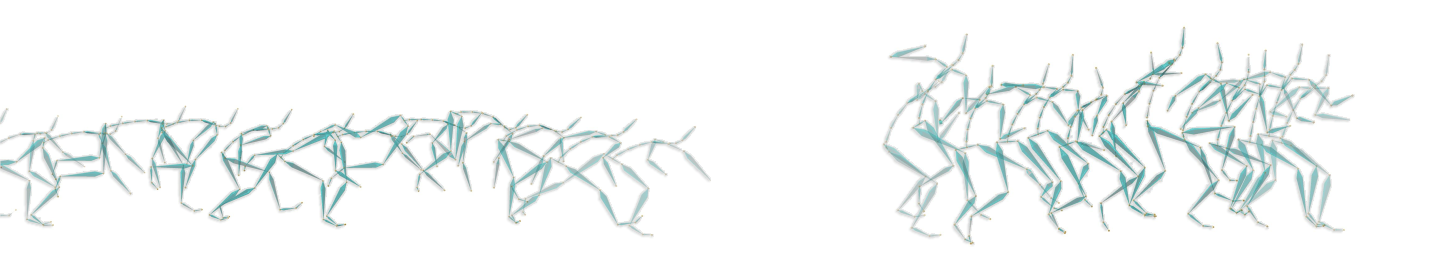}};
        \begin{scope}[x={(image.south east)},y={(image.north west)}]
            \node at (0.22, \pll) {source motion (ogre)};
            \node at (0.8, \pll) {transfered motion (clown)};
        \end{scope}
    \end{tikzpicture}
    
    \caption{Motion characterization. The walking motion of the ogre is transferred to the clown. Our method preserves the semantics of the motion, while the result motion is highly characterized.}
    \label{fig:characterization}
\end{figure*}

An immediate application of our improved motion matching can be motion stylization and characterization. We show that by training different VQ-PAEs on different characters from MOCHA~\cite{jang2023mocha} dataset in a shared phase manifold, we can transfer the core content of motion among different characters, and stylize the motion according to a specific character dataset as shown in \Cref{fig:characterization}. We are able to achieve a similar effect as the motion stylization method proposed by \citet{jang2023mocha} with a much simpler setup. Since the code for MOCHA~\cite{jang2023mocha} is not available, we provide a qualitative comparison in the accompanying video.

\subsection{Ablation study}
\label{sec:ablation_study}

We study the impact of codebook size and usage of reinitialization of $\MA$ on the performance of our method.

\paragraph{Codebook size} Choosing an appropriate codebook size is critical for our framework, as a small codebook size will not be able to capture the different semantics, and a large codebook makes the alignment on semantics less accurate. We measure the expressiveness of a learned phase manifold by calculating the mean joint position error when using MLP to reconstruct the input motion from the phase manifold embeddings, using the same setting as in \Cref{sec:motion_alignment}. Note that MOCHA datasets have a larger error due to a large number of transitions between amplitudes, which cannot be captured by the per-frame decoding MLP, but can be faithfully reconstructed by the motion matching algorithm using a sequence of embeddings as input.
As shown in \Cref{tab:rec_error}, the expressiveness reaches a plateau when the codebook size is larger than 64 for Dog and Human-Loco dataset, and 64 for MOCHA dataset, but peaks at 512. However, we also show that the percentage of embeddings in the dataset that lies on a shared connected component \textit{decreases} with the codebook size, as shown in \Cref{tab:overlapping}. This indicates that a large codebook size can cause a disparity in the learned manifold embeddings, in favor of higher expressiveness. Although size 512 improves on expressiveness, it fails to create sufficient overlapping between datasets. Thus, we choose $|\MA| = 32$ for the human-dog setting and $|\MA| = 64$ for the stylized setting in our experiments according to the results.

\begin{table}
    \centering
    \caption{Per-frame mean joint position error (cm) using MLP.}
    \small
    \begin{tabular}{lcccccc}
        \toprule
        Size of $\MA$ & 8 & 16 & 32 & 64 & 128 & 512 \\
        \midrule
        Dog~\shortcite{zhang2018mode} & 1.86 & 1.67 & 1.41 & 1.24 & 1.19 & 0.87 \\
        Human-Locomotion~\shortcite{starke2019neural} & 1.29 & 1.26 & 1.19 & 1.13 & 1.08 & 1.01 \\
        MOCHA-Clown~\shortcite{jang2023mocha} & 11.6 & 10.1 & 9.97 & 9.86 & 9.29 & 6.50 \\
        MOCHA-Ogre~\shortcite{jang2023mocha} & 12.2 & 11.3 & 10.9 & 9.85 & 9.42 & 7.70\\
        \bottomrule
    \end{tabular}
    \label{tab:rec_error}
\end{table}

\begin{table}
    \centering
    \caption{Manifold overlapping percentage.}
    \small
    \begin{tabular}{lcccccc}
        \toprule
        Size of $\MA$ & 8 & 16 & 32 & 64 & 128 & 512\\
        \midrule
        human-dog & 100 & 100 & 100 & 67.8 & 5.42 & 0.00\\
        stylized & 100 & 100 & 100 & 100 & 92.3 & 10.3 \\
        human-dog (no reinit.) & 100 & 73.6 & 64.2 & 52.5 & 1.32 & 0.00\\
        \bottomrule
    \end{tabular}
    \label{tab:overlapping}
\end{table}

\paragraph{Reinitialization of $\MA$} With the help of reinitialization adapted from \citet{zheng2023online}, every entry in $\MA$ is used by both VQ-PAEs, which is crucial for building a common phase manifold. When disabled, the phase manifold overlapping percentage drops as shown in \Cref{tab:overlapping}.
\section{Discussion and Conclusion}

In this work, we present a disconnected 1D phase manifold for motion alignment, leveraging the intrinsic periodicity of motions. We show that the alignment can be achieved thanks to the carefully designed \emph{structure} of the latent space. With the proposed vector quantized periodic autoencoder, we can embed motions from different characters with different skeletal structures or morphologies into the same phase manifold without any supervision or skeletal structure correspondences. We demonstrate that when integrated with motion matching, various applications such as motion retrieval, transfer, and stylization can be achieved.

The key success of our simple motion alignment lies in the limited capability of the shallow VQ-PAE, which prevents a large distortion between the motion representation and the latent embeddings, and the design of the compact latent space, a collection of ellipses embedded in $\R^d$. For semantic alignment, the structural similarity between motion datasets is explicitly reflected in the latent space through the amplitudes. For example, running motions are clustered into ellipses with larger amplitudes, while idling motions are clustered into ellipses with smaller amplitudes. As for timing alignment, the anisotropic structure of the ellipses (\Cref{eq:phase_manifold}) is crucial. Although we expect the phase variable to progress linearly through an entire motion cycle, the progress of the phase manifold is not linear. This guarantees, for example, that crucial points in motions such as foot contacts, are mapped to the vertices of ellipses. However, this alignment is not always perfect: as can be seen at 3:01 in the accompanying video, a mismatch of the left and right foot contact exists, since no joint correspondence is provided, so the left and right body parts are indistinguishable.

While our current framework provides good timing alignment, the semantics alignment is not always perfect. It requires carefully picking the right codebook size to balance between expressiveness and the amount of overlap among datasets. It also implicitly requires the datasets to contain semantically similar motion distributions. For example, the backward motion is presented in the Human-Loco dataset but not in the Dog dataset, so the Human backward walking is aligned with forward walking for Dog. In the future, it would be interesting to automatically learn the size of $\MA$ and filter out motions that are not semantically similar. In addition, the residual amplitude, removed by the quantization, could be potentially used for representing ``styles'' of motions within the same semantics. 

Our current framework is not generative. It would be interesting to explore the possibility of generating new motions from the phase manifold. Another promising direction for future research is training the PAEs with other 1D input signals, such as a music dataset, e.g.\ for a tightly aligned music-to-dance generation.

\begin{acks}
    We thank the anonymous reviewers for their valuable feedback. We also thank Heyuan Yao and Alexander Winkler for the insightful discussions. This work was supported in part by the ERC Consolidator Grant No. 101003104 (MYCLOTH).
\end{acks}

\bibliographystyle{ACM-Reference-Format}
\bibliography{bibs}


\begin{thebibliography}{41}


\ifx \showCODEN    \undefined \def \showCODEN     #1{\unskip}     \fi
\ifx \showDOI      \undefined \def \showDOI       #1{#1}\fi
\ifx \showISBNx    \undefined \def \showISBNx     #1{\unskip}     \fi
\ifx \showISBNxiii \undefined \def \showISBNxiii  #1{\unskip}     \fi
\ifx \showISSN     \undefined \def \showISSN      #1{\unskip}     \fi
\ifx \showLCCN     \undefined \def \showLCCN      #1{\unskip}     \fi
\ifx \shownote     \undefined \def \shownote      #1{#1}          \fi
\ifx \showarticletitle \undefined \def \showarticletitle #1{#1}   \fi
\ifx \showURL      \undefined \def \showURL       {\relax}        \fi
\providecommand\bibfield[2]{#2}
\providecommand\bibinfo[2]{#2}
\providecommand\natexlab[1]{#1}
\providecommand\showeprint[2][]{arXiv:#2}

\bibitem[\protect\citeauthoryear{Aberman, Li, Lischinski, Sorkine-Hornung, Cohen-Or, and Chen}{Aberman et~al\mbox{.}}{2020a}]%
        {aberman2020skeleton}
\bibfield{author}{\bibinfo{person}{Kfir Aberman}, \bibinfo{person}{Peizhuo Li}, \bibinfo{person}{Dani Lischinski}, \bibinfo{person}{Olga Sorkine-Hornung}, \bibinfo{person}{Daniel Cohen-Or}, {and} \bibinfo{person}{Baoquan Chen}.} \bibinfo{year}{2020}\natexlab{a}.
\newblock \showarticletitle{Skeleton-aware networks for deep motion retargeting}.
\newblock \bibinfo{journal}{\emph{ACM Transactions on Graphics (TOG)}} \bibinfo{volume}{39}, \bibinfo{number}{4} (\bibinfo{year}{2020}), \bibinfo{pages}{62--1}.
\newblock


\bibitem[\protect\citeauthoryear{Aberman, Weng, Lischinski, Cohen-Or, and Chen}{Aberman et~al\mbox{.}}{2020b}]%
        {aberman2020unpaired}
\bibfield{author}{\bibinfo{person}{Kfir Aberman}, \bibinfo{person}{Yijia Weng}, \bibinfo{person}{Dani Lischinski}, \bibinfo{person}{Daniel Cohen-Or}, {and} \bibinfo{person}{Baoquan Chen}.} \bibinfo{year}{2020}\natexlab{b}.
\newblock \showarticletitle{Unpaired motion style transfer from video to animation}.
\newblock \bibinfo{journal}{\emph{ACM Transactions on Graphics (TOG)}} \bibinfo{volume}{39}, \bibinfo{number}{4} (\bibinfo{year}{2020}), \bibinfo{pages}{64--1}.
\newblock


\bibitem[\protect\citeauthoryear{Arikan and Forsyth}{Arikan and Forsyth}{2002}]%
        {arikan2002interactive}
\bibfield{author}{\bibinfo{person}{Okan Arikan} {and} \bibinfo{person}{David~A Forsyth}.} \bibinfo{year}{2002}\natexlab{}.
\newblock \showarticletitle{Interactive motion generation from examples}.
\newblock \bibinfo{journal}{\emph{ACM Transactions on Graphics (TOG)}} \bibinfo{volume}{21}, \bibinfo{number}{3} (\bibinfo{year}{2002}), \bibinfo{pages}{483--490}.
\newblock


\bibitem[\protect\citeauthoryear{Aristidou, Cohen-Or, Hodgins, Chrysanthou, and Shamir}{Aristidou et~al\mbox{.}}{2018}]%
        {aristidou2018deep}
\bibfield{author}{\bibinfo{person}{Andreas Aristidou}, \bibinfo{person}{Daniel Cohen-Or}, \bibinfo{person}{Jessica~K Hodgins}, \bibinfo{person}{Yiorgos Chrysanthou}, {and} \bibinfo{person}{Ariel Shamir}.} \bibinfo{year}{2018}\natexlab{}.
\newblock \showarticletitle{Deep motifs and motion signatures}.
\newblock \bibinfo{journal}{\emph{ACM Transactions on Graphics (TOG)}} \bibinfo{volume}{37}, \bibinfo{number}{6} (\bibinfo{year}{2018}), \bibinfo{pages}{1--13}.
\newblock


\bibitem[\protect\citeauthoryear{Büttner and Clavet}{Büttner and Clavet}{2015}]%
        {motionmatching}
\bibfield{author}{\bibinfo{person}{Michael Büttner} {and} \bibinfo{person}{Simon Clavet}.} \bibinfo{year}{2015}\natexlab{}.
\newblock \bibinfo{title}{Motion Matching - The Road to Next Gen Animation.}
\newblock
\newblock
\urldef\tempurl%
\url{https://www.youtube.com/watch?v=z_wpgHFSWss}
\showURL{%
\tempurl}


\bibitem[\protect\citeauthoryear{Choi and Ko}{Choi and Ko}{2000}]%
        {choi2000online}
\bibfield{author}{\bibinfo{person}{Kwang-Jin Choi} {and} \bibinfo{person}{Hyeong-Seok Ko}.} \bibinfo{year}{2000}\natexlab{}.
\newblock \showarticletitle{Online motion retargetting}.
\newblock \bibinfo{journal}{\emph{The Journal of Visualization and Computer Animation}} \bibinfo{volume}{11}, \bibinfo{number}{5} (\bibinfo{year}{2000}), \bibinfo{pages}{223--235}.
\newblock


\bibitem[\protect\citeauthoryear{Dhariwal, Jun, Payne, Kim, Radford, and Sutskever}{Dhariwal et~al\mbox{.}}{2020}]%
        {dhariwal2020jukebox}
\bibfield{author}{\bibinfo{person}{Prafulla Dhariwal}, \bibinfo{person}{Heewoo Jun}, \bibinfo{person}{Christine Payne}, \bibinfo{person}{Jong~Wook Kim}, \bibinfo{person}{Alec Radford}, {and} \bibinfo{person}{Ilya Sutskever}.} \bibinfo{year}{2020}\natexlab{}.
\newblock \showarticletitle{Jukebox: A generative model for music}.
\newblock \bibinfo{journal}{\emph{arXiv preprint arXiv:2005.00341}} (\bibinfo{year}{2020}).
\newblock


\bibitem[\protect\citeauthoryear{Geng, Wang, Wei, Liu, Li, and Hu}{Geng et~al\mbox{.}}{2023}]%
        {geng2023human}
\bibfield{author}{\bibinfo{person}{Zigang Geng}, \bibinfo{person}{Chunyu Wang}, \bibinfo{person}{Yixuan Wei}, \bibinfo{person}{Ze Liu}, \bibinfo{person}{Houqiang Li}, {and} \bibinfo{person}{Han Hu}.} \bibinfo{year}{2023}\natexlab{}.
\newblock \showarticletitle{Human Pose as Compositional Tokens}. In \bibinfo{booktitle}{\emph{Proceedings of the IEEE/CVF Conference on Computer Vision and Pattern Recognition}}. \bibinfo{pages}{660--671}.
\newblock


\bibitem[\protect\citeauthoryear{Gleicher}{Gleicher}{1998}]%
        {gleicher1998retargetting}
\bibfield{author}{\bibinfo{person}{Michael Gleicher}.} \bibinfo{year}{1998}\natexlab{}.
\newblock \showarticletitle{Retargetting motion to new characters}. In \bibinfo{booktitle}{\emph{Proc.~25th annual conference on computer graphics and interactive techniques}}. ACM, \bibinfo{pages}{33--42}.
\newblock


\bibitem[\protect\citeauthoryear{Goodfellow, Pouget-Abadie, Mirza, Xu, Warde-Farley, Ozair, Courville, and Bengio}{Goodfellow et~al\mbox{.}}{2020}]%
        {goodfellow2020generative}
\bibfield{author}{\bibinfo{person}{Ian Goodfellow}, \bibinfo{person}{Jean Pouget-Abadie}, \bibinfo{person}{Mehdi Mirza}, \bibinfo{person}{Bing Xu}, \bibinfo{person}{David Warde-Farley}, \bibinfo{person}{Sherjil Ozair}, \bibinfo{person}{Aaron Courville}, {and} \bibinfo{person}{Yoshua Bengio}.} \bibinfo{year}{2020}\natexlab{}.
\newblock \showarticletitle{Generative adversarial networks}.
\newblock \bibinfo{journal}{\emph{Commun. ACM}} \bibinfo{volume}{63}, \bibinfo{number}{11} (\bibinfo{year}{2020}), \bibinfo{pages}{139--144}.
\newblock


\bibitem[\protect\citeauthoryear{Guo, Zuo, Wang, and Cheng}{Guo et~al\mbox{.}}{2022}]%
        {guo2022tm2t}
\bibfield{author}{\bibinfo{person}{Chuan Guo}, \bibinfo{person}{Xinxin Zuo}, \bibinfo{person}{Sen Wang}, {and} \bibinfo{person}{Li Cheng}.} \bibinfo{year}{2022}\natexlab{}.
\newblock \showarticletitle{Tm2t: Stochastic and tokenized modeling for the reciprocal generation of 3d human motions and texts}. In \bibinfo{booktitle}{\emph{European Conference on Computer Vision}}. Springer, \bibinfo{pages}{580--597}.
\newblock


\bibitem[\protect\citeauthoryear{Holden, Komura, and Saito}{Holden et~al\mbox{.}}{2017}]%
        {holden2017phase}
\bibfield{author}{\bibinfo{person}{Daniel Holden}, \bibinfo{person}{Taku Komura}, {and} \bibinfo{person}{Jun Saito}.} \bibinfo{year}{2017}\natexlab{}.
\newblock \showarticletitle{Phase-functioned neural networks for character control}.
\newblock \bibinfo{journal}{\emph{ACM Transactions on Graphics (TOG)}} \bibinfo{volume}{36}, \bibinfo{number}{4} (\bibinfo{year}{2017}), \bibinfo{pages}{1--13}.
\newblock


\bibitem[\protect\citeauthoryear{Jang, Ye, Won, and Lee}{Jang et~al\mbox{.}}{2023}]%
        {jang2023mocha}
\bibfield{author}{\bibinfo{person}{Deok-Kyeong Jang}, \bibinfo{person}{Yuting Ye}, \bibinfo{person}{Jungdam Won}, {and} \bibinfo{person}{Sung-Hee Lee}.} \bibinfo{year}{2023}\natexlab{}.
\newblock \showarticletitle{MOCHA: Real-Time Motion Characterization via Context Matching}. In \bibinfo{booktitle}{\emph{SIGGRAPH Asia 2023 Conference Papers}}. \bibinfo{pages}{1--11}.
\newblock


\bibitem[\protect\citeauthoryear{Kim, Xie, and van~de Panne}{Kim et~al\mbox{.}}{2020}]%
        {kim2020learning}
\bibfield{author}{\bibinfo{person}{Nam~Hee Kim}, \bibinfo{person}{Zhaoming Xie}, {and} \bibinfo{person}{Michiel van~de Panne}.} \bibinfo{year}{2020}\natexlab{}.
\newblock \showarticletitle{Learning to Correspond Dynamical Systems}. In \bibinfo{booktitle}{\emph{Proceedings of the 2nd Conference on Learning for Dynamics and Control}} \emph{(\bibinfo{series}{Proceedings of Machine Learning Research})}, \bibfield{editor}{\bibinfo{person}{Alexandre~M. Bayen}, \bibinfo{person}{Ali Jadbabaie}, \bibinfo{person}{George Pappas}, \bibinfo{person}{Pablo~A. Parrilo}, \bibinfo{person}{Benjamin Recht}, \bibinfo{person}{Claire Tomlin}, {and} \bibinfo{person}{Melanie Zeilinger}} (Eds.), Vol.~\bibinfo{volume}{120}. \bibinfo{publisher}{PMLR}, \bibinfo{pages}{105--117}.
\newblock
\urldef\tempurl%
\url{https://proceedings.mlr.press/v120/kim20a.html}
\showURL{%
\tempurl}


\bibitem[\protect\citeauthoryear{Kim, Sorokin, Lee, and Ha}{Kim et~al\mbox{.}}{2022}]%
        {kim2022humanconquad}
\bibfield{author}{\bibinfo{person}{Sunwoo Kim}, \bibinfo{person}{Maks Sorokin}, \bibinfo{person}{Jehee Lee}, {and} \bibinfo{person}{Sehoon Ha}.} \bibinfo{year}{2022}\natexlab{}.
\newblock \showarticletitle{Humanconquad: human motion control of quadrupedal robots using deep reinforcement learning}.
\newblock In \bibinfo{booktitle}{\emph{SIGGRAPH Asia 2022 Emerging Technologies}}. \bibinfo{pages}{1--2}.
\newblock


\bibitem[\protect\citeauthoryear{Kovar, Gleicher, and Pighin}{Kovar et~al\mbox{.}}{2002}]%
        {kovar2002motion}
\bibfield{author}{\bibinfo{person}{Lucas Kovar}, \bibinfo{person}{Michael Gleicher}, {and} \bibinfo{person}{Fr\'{e}d\'{e}ric Pighin}.} \bibinfo{year}{2002}\natexlab{}.
\newblock \showarticletitle{Motion Graphs}. In \bibinfo{booktitle}{\emph{Proceedings of the 29th Annual Conference on Computer Graphics and Interactive Techniques}} \emph{(\bibinfo{series}{SIGGRAPH '02})}. \bibinfo{publisher}{Association for Computing Machinery}, \bibinfo{address}{New York, NY, USA}, \bibinfo{pages}{473–482}.
\newblock
\showISBNx{1581135211}
\urldef\tempurl%
\url{https://doi.org/10.1145/566570.566605}
\showDOI{\tempurl}


\bibitem[\protect\citeauthoryear{Lee and Shin}{Lee and Shin}{1999}]%
        {lee1999hierarchical}
\bibfield{author}{\bibinfo{person}{Jehee Lee} {and} \bibinfo{person}{Sung~Yong Shin}.} \bibinfo{year}{1999}\natexlab{}.
\newblock \showarticletitle{A hierarchical approach to interactive motion editing for human-like figures}. In \bibinfo{booktitle}{\emph{Proc.~26th annual conference on computer graphics and interactive techniques}}. ACM Press/Addison-Wesley Publishing Co., \bibinfo{pages}{39--48}.
\newblock


\bibitem[\protect\citeauthoryear{Lee, Wampler, Bernstein, Popovi{\'c}, and Popovi{\'c}}{Lee et~al\mbox{.}}{2010}]%
        {motionfield}
\bibfield{author}{\bibinfo{person}{Yongjoon Lee}, \bibinfo{person}{Kevin Wampler}, \bibinfo{person}{Gilbert Bernstein}, \bibinfo{person}{Jovan Popovi{\'c}}, {and} \bibinfo{person}{Zoran Popovi{\'c}}.} \bibinfo{year}{2010}\natexlab{}.
\newblock \showarticletitle{Motion fields for interactive character locomotion}.
\newblock \bibinfo{pages}{1--8}.
\newblock


\bibitem[\protect\citeauthoryear{Li, Won, Clegg, Kim, Rai, and Ha}{Li et~al\mbox{.}}{2023}]%
        {li2023ace}
\bibfield{author}{\bibinfo{person}{Tianyu Li}, \bibinfo{person}{Jungdam Won}, \bibinfo{person}{Alexander Clegg}, \bibinfo{person}{Jeonghwan Kim}, \bibinfo{person}{Akshara Rai}, {and} \bibinfo{person}{Sehoon Ha}.} \bibinfo{year}{2023}\natexlab{}.
\newblock \showarticletitle{Ace: Adversarial correspondence embedding for cross morphology motion retargeting from human to nonhuman characters}. In \bibinfo{booktitle}{\emph{SIGGRAPH Asia 2023 Conference Papers}}. \bibinfo{pages}{1--11}.
\newblock


\bibitem[\protect\citeauthoryear{Lim, Chang, and Choi}{Lim et~al\mbox{.}}{2019}]%
        {lim2019pmnet}
\bibfield{author}{\bibinfo{person}{Jongin Lim}, \bibinfo{person}{Hyung~Jin Chang}, {and} \bibinfo{person}{Jin~Young Choi}.} \bibinfo{year}{2019}\natexlab{}.
\newblock \showarticletitle{PMnet: Learning of Disentangled Pose and Movement for Unsupervised Motion Retargeting.}. In \bibinfo{booktitle}{\emph{BMVC}}, Vol.~\bibinfo{volume}{2}. \bibinfo{pages}{7}.
\newblock


\bibitem[\protect\citeauthoryear{Mason}{Mason}{2022}]%
        {IanWebPost}
\bibfield{author}{\bibinfo{person}{Ian Mason}.} \bibinfo{year}{2022}\natexlab{}.
\newblock \bibinfo{title}{Periodic Autoencoder - Explanation and Addendum}.
\newblock
\newblock
\urldef\tempurl%
\url{https://www.ianxmason.com/posts/PAE/}
\showURL{%
\tempurl}


\bibitem[\protect\citeauthoryear{Min and Chai}{Min and Chai}{2012}]%
        {min2012motion}
\bibfield{author}{\bibinfo{person}{Jianyuan Min} {and} \bibinfo{person}{Jinxiang Chai}.} \bibinfo{year}{2012}\natexlab{}.
\newblock \showarticletitle{Motion graphs++ a compact generative model for semantic motion analysis and synthesis}.
\newblock \bibinfo{journal}{\emph{ACM Transactions on Graphics (TOG)}} \bibinfo{volume}{31}, \bibinfo{number}{6} (\bibinfo{year}{2012}), \bibinfo{pages}{1--12}.
\newblock


\bibitem[\protect\citeauthoryear{Park, Lee, and Lee}{Park et~al\mbox{.}}{2011}]%
        {park2011finding}
\bibfield{author}{\bibinfo{person}{Jong~Pil Park}, \bibinfo{person}{Kang~Hoon Lee}, {and} \bibinfo{person}{Jehee Lee}.} \bibinfo{year}{2011}\natexlab{}.
\newblock \showarticletitle{Finding syntactic structures from human motion data}. In \bibinfo{booktitle}{\emph{Computer Graphics Forum}}, Vol.~\bibinfo{volume}{30}. Wiley Online Library, \bibinfo{pages}{2183--2193}.
\newblock


\bibitem[\protect\citeauthoryear{Park, Shin, and Shin}{Park et~al\mbox{.}}{2002}]%
        {park2002line}
\bibfield{author}{\bibinfo{person}{Sang~Il Park}, \bibinfo{person}{Hyun~Joon Shin}, {and} \bibinfo{person}{Sung~Yong Shin}.} \bibinfo{year}{2002}\natexlab{}.
\newblock \showarticletitle{On-line locomotion generation based on motion blending}. In \bibinfo{booktitle}{\emph{Proceedings of the 2002 ACM SIGGRAPH/Eurographics symposium on Computer animation}}. \bibinfo{pages}{105--111}.
\newblock


\bibitem[\protect\citeauthoryear{Peng, Abbeel, Levine, and Van~de Panne}{Peng et~al\mbox{.}}{2018}]%
        {peng2018deepmimic}
\bibfield{author}{\bibinfo{person}{Xue~Bin Peng}, \bibinfo{person}{Pieter Abbeel}, \bibinfo{person}{Sergey Levine}, {and} \bibinfo{person}{Michiel Van~de Panne}.} \bibinfo{year}{2018}\natexlab{}.
\newblock \showarticletitle{Deepmimic: Example-guided deep reinforcement learning of physics-based character skills}.
\newblock \bibinfo{journal}{\emph{ACM Transactions On Graphics (TOG)}} \bibinfo{volume}{37}, \bibinfo{number}{4} (\bibinfo{year}{2018}), \bibinfo{pages}{1--14}.
\newblock


\bibitem[\protect\citeauthoryear{Rombach, Blattmann, Lorenz, Esser, and Ommer}{Rombach et~al\mbox{.}}{2022}]%
        {rombach2022high}
\bibfield{author}{\bibinfo{person}{Robin Rombach}, \bibinfo{person}{Andreas Blattmann}, \bibinfo{person}{Dominik Lorenz}, \bibinfo{person}{Patrick Esser}, {and} \bibinfo{person}{Bj{\"o}rn Ommer}.} \bibinfo{year}{2022}\natexlab{}.
\newblock \showarticletitle{High-resolution image synthesis with latent diffusion models}. In \bibinfo{booktitle}{\emph{Proceedings of the IEEE/CVF conference on computer vision and pattern recognition}}. \bibinfo{pages}{10684--10695}.
\newblock


\bibitem[\protect\citeauthoryear{Shi, Starke, Ye, Komura, and Won}{Shi et~al\mbox{.}}{2023}]%
        {shi2023phasemp}
\bibfield{author}{\bibinfo{person}{Mingyi Shi}, \bibinfo{person}{Sebastian Starke}, \bibinfo{person}{Yuting Ye}, \bibinfo{person}{Taku Komura}, {and} \bibinfo{person}{Jungdam Won}.} \bibinfo{year}{2023}\natexlab{}.
\newblock \showarticletitle{Phasemp: Robust 3d pose estimation via phase-conditioned human motion prior}. In \bibinfo{booktitle}{\emph{Proceedings of the IEEE/CVF International Conference on Computer Vision}}. \bibinfo{pages}{14725--14737}.
\newblock


\bibitem[\protect\citeauthoryear{Siyao, Yu, Gu, Lin, Wang, Qian, Loy, and Liu}{Siyao et~al\mbox{.}}{2022}]%
        {siyao2022bailando}
\bibfield{author}{\bibinfo{person}{Li Siyao}, \bibinfo{person}{Weijiang Yu}, \bibinfo{person}{Tianpei Gu}, \bibinfo{person}{Chunze Lin}, \bibinfo{person}{Quan Wang}, \bibinfo{person}{Chen Qian}, \bibinfo{person}{Chen~Change Loy}, {and} \bibinfo{person}{Ziwei Liu}.} \bibinfo{year}{2022}\natexlab{}.
\newblock \showarticletitle{Bailando: 3d dance generation by actor-critic gpt with choreographic memory}. In \bibinfo{booktitle}{\emph{Proceedings of the IEEE/CVF Conference on Computer Vision and Pattern Recognition}}. \bibinfo{pages}{11050--11059}.
\newblock


\bibitem[\protect\citeauthoryear{Starke, Starke, Komura, and Steinicke}{Starke et~al\mbox{.}}{2023}]%
        {starke2023phasebetweener}
\bibfield{author}{\bibinfo{person}{Paul Starke}, \bibinfo{person}{Sebastian Starke}, \bibinfo{person}{Taku Komura}, {and} \bibinfo{person}{Frank Steinicke}.} \bibinfo{year}{2023}\natexlab{}.
\newblock \showarticletitle{Motion In-Betweening with Phase Manifolds}.
\newblock \bibinfo{journal}{\emph{Proceedings of the ACM on Computer Graphics and Interactive Techniques}} \bibinfo{volume}{6}, \bibinfo{number}{3} (\bibinfo{date}{Aug.} \bibinfo{year}{2023}), \bibinfo{pages}{1–17}.
\newblock
\showISSN{2577-6193}
\urldef\tempurl%
\url{https://doi.org/10.1145/3606921}
\showDOI{\tempurl}


\bibitem[\protect\citeauthoryear{Starke, Mason, and Komura}{Starke et~al\mbox{.}}{2022}]%
        {starke2022deepphase}
\bibfield{author}{\bibinfo{person}{Sebastian Starke}, \bibinfo{person}{Ian Mason}, {and} \bibinfo{person}{Taku Komura}.} \bibinfo{year}{2022}\natexlab{}.
\newblock \showarticletitle{Deepphase: Periodic autoencoders for learning motion phase manifolds}.
\newblock \bibinfo{journal}{\emph{ACM Transactions on Graphics (TOG)}} \bibinfo{volume}{41}, \bibinfo{number}{4} (\bibinfo{year}{2022}), \bibinfo{pages}{1--13}.
\newblock


\bibitem[\protect\citeauthoryear{Starke, Zhang, Komura, and Saito}{Starke et~al\mbox{.}}{2019}]%
        {starke2019neural}
\bibfield{author}{\bibinfo{person}{Sebastian Starke}, \bibinfo{person}{He Zhang}, \bibinfo{person}{Taku Komura}, {and} \bibinfo{person}{Jun Saito}.} \bibinfo{year}{2019}\natexlab{}.
\newblock \showarticletitle{Neural state machine for character-scene interactions.}
\newblock \bibinfo{journal}{\emph{ACM Trans. Graph.}} \bibinfo{volume}{38}, \bibinfo{number}{6} (\bibinfo{year}{2019}), \bibinfo{pages}{209--1}.
\newblock


\bibitem[\protect\citeauthoryear{Starke, Zhao, Komura, and Zaman}{Starke et~al\mbox{.}}{2020}]%
        {starke2020local}
\bibfield{author}{\bibinfo{person}{Sebastian Starke}, \bibinfo{person}{Yiwei Zhao}, \bibinfo{person}{Taku Komura}, {and} \bibinfo{person}{Kazi Zaman}.} \bibinfo{year}{2020}\natexlab{}.
\newblock \showarticletitle{Local motion phases for learning multi-contact character movements}.
\newblock \bibinfo{journal}{\emph{ACM Transactions on Graphics (TOG)}} \bibinfo{volume}{39}, \bibinfo{number}{4} (\bibinfo{year}{2020}), \bibinfo{pages}{54--1}.
\newblock


\bibitem[\protect\citeauthoryear{Tak and Ko}{Tak and Ko}{2005}]%
        {tak2005physically}
\bibfield{author}{\bibinfo{person}{Seyoon Tak} {and} \bibinfo{person}{Hyeong-Seok Ko}.} \bibinfo{year}{2005}\natexlab{}.
\newblock \showarticletitle{A physically-based motion retargeting filter}.
\newblock \bibinfo{journal}{\emph{ACM Trans.~Graph.}} \bibinfo{volume}{24}, \bibinfo{number}{1} (\bibinfo{year}{2005}), \bibinfo{pages}{98--117}.
\newblock


\bibitem[\protect\citeauthoryear{Unuma, Anjyo, and Takeuchi}{Unuma et~al\mbox{.}}{1995}]%
        {unuma1995fourier}
\bibfield{author}{\bibinfo{person}{Munetoshi Unuma}, \bibinfo{person}{Ken Anjyo}, {and} \bibinfo{person}{Ryozo Takeuchi}.} \bibinfo{year}{1995}\natexlab{}.
\newblock \showarticletitle{Fourier principles for emotion-based human figure animation}. In \bibinfo{booktitle}{\emph{Proceedings of the 22nd annual conference on Computer graphics and interactive techniques}}. \bibinfo{pages}{91--96}.
\newblock


\bibitem[\protect\citeauthoryear{Van Den~Oord, Vinyals, et~al\mbox{.}}{Van Den~Oord et~al\mbox{.}}{2017}]%
        {van2017neural}
\bibfield{author}{\bibinfo{person}{Aaron Van Den~Oord}, \bibinfo{person}{Oriol Vinyals}, {et~al\mbox{.}}} \bibinfo{year}{2017}\natexlab{}.
\newblock \showarticletitle{Neural discrete representation learning}.
\newblock \bibinfo{journal}{\emph{Advances in neural information processing systems}}  \bibinfo{volume}{30} (\bibinfo{year}{2017}).
\newblock


\bibitem[\protect\citeauthoryear{Villegas, Yang, Ceylan, and Lee}{Villegas et~al\mbox{.}}{2018}]%
        {villegas2018neural}
\bibfield{author}{\bibinfo{person}{Ruben Villegas}, \bibinfo{person}{Jimei Yang}, \bibinfo{person}{Duygu Ceylan}, {and} \bibinfo{person}{Honglak Lee}.} \bibinfo{year}{2018}\natexlab{}.
\newblock \showarticletitle{Neural Kinematic Networks for Unsupervised Motion Retargetting}. In \bibinfo{booktitle}{\emph{Proc.~IEEE CVPR}}. \bibinfo{pages}{8639--8648}.
\newblock


\bibitem[\protect\citeauthoryear{Xia, Wang, Chai, and Hodgins}{Xia et~al\mbox{.}}{2015}]%
        {xia2015realtime}
\bibfield{author}{\bibinfo{person}{Shihong Xia}, \bibinfo{person}{Congyi Wang}, \bibinfo{person}{Jinxiang Chai}, {and} \bibinfo{person}{Jessica Hodgins}.} \bibinfo{year}{2015}\natexlab{}.
\newblock \showarticletitle{Realtime style transfer for unlabeled heterogeneous human motion}.
\newblock \bibinfo{journal}{\emph{ACM Transactions on Graphics (TOG)}} \bibinfo{volume}{34}, \bibinfo{number}{4} (\bibinfo{year}{2015}), \bibinfo{pages}{1--10}.
\newblock


\bibitem[\protect\citeauthoryear{Zhang, Starke, Komura, and Saito}{Zhang et~al\mbox{.}}{2018}]%
        {zhang2018mode}
\bibfield{author}{\bibinfo{person}{He Zhang}, \bibinfo{person}{Sebastian Starke}, \bibinfo{person}{Taku Komura}, {and} \bibinfo{person}{Jun Saito}.} \bibinfo{year}{2018}\natexlab{}.
\newblock \showarticletitle{Mode-adaptive neural networks for quadruped motion control}.
\newblock \bibinfo{journal}{\emph{ACM Transactions on Graphics (TOG)}} \bibinfo{volume}{37}, \bibinfo{number}{4} (\bibinfo{year}{2018}), \bibinfo{pages}{1--11}.
\newblock


\bibitem[\protect\citeauthoryear{Zhang, Zhang, Cun, Huang, Zhang, Zhao, Lu, and Shen}{Zhang et~al\mbox{.}}{2023}]%
        {zhang2023t2m}
\bibfield{author}{\bibinfo{person}{Jianrong Zhang}, \bibinfo{person}{Yangsong Zhang}, \bibinfo{person}{Xiaodong Cun}, \bibinfo{person}{Shaoli Huang}, \bibinfo{person}{Yong Zhang}, \bibinfo{person}{Hongwei Zhao}, \bibinfo{person}{Hongtao Lu}, {and} \bibinfo{person}{Xi Shen}.} \bibinfo{year}{2023}\natexlab{}.
\newblock \showarticletitle{T2m-gpt: Generating human motion from textual descriptions with discrete representations}.
\newblock \bibinfo{journal}{\emph{arXiv preprint arXiv:2301.06052}} (\bibinfo{year}{2023}).
\newblock


\bibitem[\protect\citeauthoryear{Zheng and Vedaldi}{Zheng and Vedaldi}{2023}]%
        {zheng2023online}
\bibfield{author}{\bibinfo{person}{Chuanxia Zheng} {and} \bibinfo{person}{Andrea Vedaldi}.} \bibinfo{year}{2023}\natexlab{}.
\newblock \showarticletitle{Online clustered codebook}. In \bibinfo{booktitle}{\emph{Proceedings of the IEEE/CVF International Conference on Computer Vision}}. \bibinfo{pages}{22798--22807}.
\newblock


\bibitem[\protect\citeauthoryear{Zhu, Park, Isola, and Efros}{Zhu et~al\mbox{.}}{2017}]%
        {zhu2017unpaired}
\bibfield{author}{\bibinfo{person}{Jun-Yan Zhu}, \bibinfo{person}{Taesung Park}, \bibinfo{person}{Phillip Isola}, {and} \bibinfo{person}{Alexei~A Efros}.} \bibinfo{year}{2017}\natexlab{}.
\newblock \showarticletitle{Unpaired image-to-image translation using cycle-consistent adversarial networks}. In \bibinfo{booktitle}{\emph{Proceedings of the IEEE international conference on computer vision}}. \bibinfo{pages}{2223--2232}.
\newblock


\end{thebibliography}



\begin{thebibliography}{7}


\ifx \showCODEN    \undefined \def \showCODEN     #1{\unskip}     \fi
\ifx \showDOI      \undefined \def \showDOI       #1{#1}\fi
\ifx \showISBNx    \undefined \def \showISBNx     #1{\unskip}     \fi
\ifx \showISBNxiii \undefined \def \showISBNxiii  #1{\unskip}     \fi
\ifx \showISSN     \undefined \def \showISSN      #1{\unskip}     \fi
\ifx \showLCCN     \undefined \def \showLCCN      #1{\unskip}     \fi
\ifx \shownote     \undefined \def \shownote      #1{#1}          \fi
\ifx \showarticletitle \undefined \def \showarticletitle #1{#1}   \fi
\ifx \showURL      \undefined \def \showURL       {\relax}        \fi
\providecommand\bibfield[2]{#2}
\providecommand\bibinfo[2]{#2}
\providecommand\natexlab[1]{#1}
\providecommand\showeprint[2][]{arXiv:#2}

\bibitem[\protect\citeauthoryear{Aberman, Li, Lischinski, Sorkine-Hornung,
  Cohen-Or, and Chen}{Aberman et~al\mbox{.}}{2020}]%
        {aberman2020skeleton}
\bibfield{author}{\bibinfo{person}{Kfir Aberman}, \bibinfo{person}{Peizhuo Li},
  \bibinfo{person}{Dani Lischinski}, \bibinfo{person}{Olga Sorkine-Hornung},
  \bibinfo{person}{Daniel Cohen-Or}, {and} \bibinfo{person}{Baoquan Chen}.}
  \bibinfo{year}{2020}\natexlab{}.
\newblock \showarticletitle{Skeleton-aware networks for deep motion
  retargeting}.
\newblock \bibinfo{journal}{\emph{ACM Transactions on Graphics (TOG)}}
  \bibinfo{volume}{39}, \bibinfo{number}{4} (\bibinfo{year}{2020}),
  \bibinfo{pages}{62--1}.
\newblock


\bibitem[\protect\citeauthoryear{Jang, Ye, Won, and Lee}{Jang
  et~al\mbox{.}}{2023}]%
        {jang2023mocha}
\bibfield{author}{\bibinfo{person}{Deok-Kyeong Jang}, \bibinfo{person}{Yuting
  Ye}, \bibinfo{person}{Jungdam Won}, {and} \bibinfo{person}{Sung-Hee Lee}.}
  \bibinfo{year}{2023}\natexlab{}.
\newblock \showarticletitle{MOCHA: Real-Time Motion Characterization via
  Context Matching}. In \bibinfo{booktitle}{\emph{SIGGRAPH Asia 2023 Conference
  Papers}}. \bibinfo{pages}{1--11}.
\newblock


\bibitem[\protect\citeauthoryear{Kingma and Ba}{Kingma and Ba}{2014}]%
        {kingma2014adam}
\bibfield{author}{\bibinfo{person}{Diederik~P Kingma} {and}
  \bibinfo{person}{Jimmy Ba}.} \bibinfo{year}{2014}\natexlab{}.
\newblock \showarticletitle{Adam: A method for stochastic optimization}.
\newblock \bibinfo{journal}{\emph{arXiv preprint arXiv:1412.6980}}
  (\bibinfo{year}{2014}).
\newblock


\bibitem[\protect\citeauthoryear{Mason}{Mason}{2022}]%
        {IanWebPost}
\bibfield{author}{\bibinfo{person}{Ian Mason}.}
  \bibinfo{year}{2022}\natexlab{}.
\newblock \bibinfo{title}{Periodic Autoencoder - Explanation and Addendum}.
\newblock
\newblock
\urldef\tempurl%
\url{https://www.ianxmason.com/posts/PAE/}
\showURL{%
\tempurl}


\bibitem[\protect\citeauthoryear{Paszke, Gross, Massa, Lerer, Bradbury, Chanan,
  Killeen, Lin, Gimelshein, Antiga, Desmaison, Kopf, Yang, DeVito, Raison,
  Tejani, Chilamkurthy, Steiner, Fang, Bai, and Chintala}{Paszke
  et~al\mbox{.}}{2019}]%
        {NEURIPS2019_9015}
\bibfield{author}{\bibinfo{person}{Adam Paszke}, \bibinfo{person}{Sam Gross},
  \bibinfo{person}{Francisco Massa}, \bibinfo{person}{Adam Lerer},
  \bibinfo{person}{James Bradbury}, \bibinfo{person}{Gregory Chanan},
  \bibinfo{person}{Trevor Killeen}, \bibinfo{person}{Zeming Lin},
  \bibinfo{person}{Natalia Gimelshein}, \bibinfo{person}{Luca Antiga},
  \bibinfo{person}{Alban Desmaison}, \bibinfo{person}{Andreas Kopf},
  \bibinfo{person}{Edward Yang}, \bibinfo{person}{Zachary DeVito},
  \bibinfo{person}{Martin Raison}, \bibinfo{person}{Alykhan Tejani},
  \bibinfo{person}{Sasank Chilamkurthy}, \bibinfo{person}{Benoit Steiner},
  \bibinfo{person}{Lu Fang}, \bibinfo{person}{Junjie Bai}, {and}
  \bibinfo{person}{Soumith Chintala}.} \bibinfo{year}{2019}\natexlab{}.
\newblock \showarticletitle{PyTorch: An Imperative Style, High-Performance Deep
  Learning Library}.
\newblock In \bibinfo{booktitle}{\emph{Advances in Neural Information
  Processing Systems 32}}, \bibfield{editor}{\bibinfo{person}{H.~Wallach},
  \bibinfo{person}{H.~Larochelle}, \bibinfo{person}{A.~Beygelzimer},
  \bibinfo{person}{F.~d\textquotesingle Alch\'{e}-Buc},
  \bibinfo{person}{E.~Fox}, {and} \bibinfo{person}{R.~Garnett}} (Eds.).
  \bibinfo{publisher}{Curran Associates, Inc.}, \bibinfo{pages}{8024--8035}.
\newblock
\urldef\tempurl%
\url{http://papers.neurips.cc/paper/9015-pytorch-an-imperative-style-high-performance-deep-learning-library.pdf}
\showURL{%
\tempurl}


\bibitem[\protect\citeauthoryear{Starke, Zhang, Komura, and Saito}{Starke
  et~al\mbox{.}}{2019}]%
        {starke2019neural}
\bibfield{author}{\bibinfo{person}{Sebastian Starke}, \bibinfo{person}{He
  Zhang}, \bibinfo{person}{Taku Komura}, {and} \bibinfo{person}{Jun Saito}.}
  \bibinfo{year}{2019}\natexlab{}.
\newblock \showarticletitle{Neural state machine for character-scene
  interactions.}
\newblock \bibinfo{journal}{\emph{ACM Trans. Graph.}} \bibinfo{volume}{38},
  \bibinfo{number}{6} (\bibinfo{year}{2019}), \bibinfo{pages}{209--1}.
\newblock


\bibitem[\protect\citeauthoryear{Zhang, Starke, Komura, and Saito}{Zhang
  et~al\mbox{.}}{2018}]%
        {zhang2018mode}
\bibfield{author}{\bibinfo{person}{He Zhang}, \bibinfo{person}{Sebastian
  Starke}, \bibinfo{person}{Taku Komura}, {and} \bibinfo{person}{Jun Saito}.}
  \bibinfo{year}{2018}\natexlab{}.
\newblock \showarticletitle{Mode-adaptive neural networks for quadruped motion
  control}.
\newblock \bibinfo{journal}{\emph{ACM Transactions on Graphics (TOG)}}
  \bibinfo{volume}{37}, \bibinfo{number}{4} (\bibinfo{year}{2018}),
  \bibinfo{pages}{1--11}.
\newblock


\end{thebibliography}

\clearpage

\clearpage

\end{document}


\title{Supplementary material: \\
WalkTheDog: Cross-Morphology Motion Alignment via Phase Manifolds}

\author{Peizhuo Li}
\orcid{0000-0001-9309-9967}
\affiliation{\institution{ETH Zurich} \country{Switzerland}}
\email{peizhuo.li@inf.ethz.ch}

\author{Sebastian Starke}
\orcid{0000-0002-4519-4326}
\affiliation{\institution{Meta Reality Labs} \country{United Kindom}}
\email{sstarke@meta.com}

\author{Yuting Ye}
\orcid{0000-0003-2643-7457}
\affiliation{\institution{Meta Reality Labs} \country{USA}}
\email{yuting.ye@meta.com}

\author{Olga Sorkine-Hornung}
\orcid{0000-0002-8089-3974}
\affiliation{\institution{ETH Zurich} \country{Switzerland}}
\email{sorkine@inf.ethz.ch}

\maketitle

\section{Implementation Details}

In this section, we give a detailed description of the phase calculation module and the motion matching (without frequency scaling) algorithm.

\subsection{Phase Calculation}

We use the equations presented by~\citet{IanWebPost} to calculate the phase $\phi$ using the entries $y_i$ from 1-channel signal $\bf Y$ and the relative timing $t_i$ from $\mathcal{T}$: 
\begin{equation}
    \label{eq:phase_calculation}
    \begin{aligned}
        s_x &= \sum_{i=1}^T y_i \cos(2\pi f t_i), \\
        s_y &= \sum_{i=1}^T y_i \sin(2\pi f t_i), \\
        \phi &= \frac{\text{atan2} (s_y, s_x)}{2\pi}, \\
    \end{aligned}
\end{equation}
where $\text{atan2} (y, x)$ is the argument of the complex number $x + i y$. This equation helps us avoid dealing with the fact that $\phi$ is not a continuous parameterization of the phase manifold.

\subsection{Motion Matching on Phase Manifolds}

Given the phase embedding sequence $\Pp$ of an input motion sequence with $T$ frames, we use \Cref{alg:mm} to retrieve the motion sequence from the database, using phase embedding as the control signal in the classical motion matching algorithm. We denote the length of replay after each match with $T_0$, the phase sequences of the database with $\Q$, the pose descriptor used to measure the similarity between frames with $\J$ and the pose with $\Y$.

\begin{algorithm}
    \caption{Phase Manifold Motion Matching}
    \label{alg:mm}
    \begin{algorithmic}
        \State $i \gets 1$
        \State $\J_{\text{start}} \gets$ initial pose descriptor
        \While {$i < T$} 
            \State $k = \argmin_{k} \|\Pp_{i:i+T_0} - \Q_{k:k+T_0}\|^2_2 + \lambda \|\J_{\text{start}} - \J_k\|^2_2$
            \State \emph{Output} $\Y_{k:k+T_0}$
            \State $i \gets i + T_0$
            \State $\J_{\text{start}} \gets \J_{k + T_0}$
        \EndWhile
    \end{algorithmic}
\end{algorithm}

In our experiment, we use a simple setup where normalized joint positions are chosen as our pose descriptor $\J$ to ensure a smooth transition between different replays. We also apply inertialization at each transition. The cost function in \Cref{alg:mm} can be customized depending on the exact application.

Since skeleton-aware networks~\cite{aberman2020skeleton} require homeomorphic skeletons, we remove the tail of the dog skeleton when compared with it. In addition, we specify a correspondence between the forelegs and arms, and the hindlegs and legs. SAN heavily requires end-effector velocity consistency between the source and target characters and struggles to transfer motion with a large difference in skeletal structure.

\subsection{Datasets}

A detailed description of datasets is listed in \Cref{tab:dataset_details}. The training time on Dog and Human-Locomotion is around 40 minutes while on MOCHA-Clown and MOCHA-Ogre is around 2 hours, proportional to the number of frames. When training on all the datasets together, it takes around 2 hours and 40 minutes.

\begin{table}
    \centering
    \caption{Details of the datasets used in our experiments.}
    \small
    \begin{tabular}{lcc}
        \toprule
        Name & Framerate & \# of Frames \\
        \midrule
        Dog~\shortcite{zhang2018mode} & 60 & 151k \\
        Human-Locomotion~\shortcite{starke2019neural} & 60 & 186k \\
        MOCHA-Clown~\shortcite{jang2023mocha} & 120 & 486k \\
        MOCHA-Ogre~\shortcite{jang2023mocha} & 120 & 500k \\
        MOCHA-Princess~\shortcite{jang2023mocha} & 120 & 501k \\
        \bottomrule
    \end{tabular}
    \label{tab:dataset_details}
\end{table}

\section{Network Architecture and Hyperparameters}

In this section, we describe the network architecture and the hyper-parameters used to train the network.

\subsection{Architecture Details}

The convolutional encoder and decoder share the same architecture of two-layer 1D convolutions with kernel size 23 with ELU as activation. The MLP producing raw amplitude $\tilde\A$ has 5 layers and the hidden units are of the same size as the amplitude. The MLP in the phase calculation module also has 5 layers. The hidden units share the size as the input frequency bin powers produced by FFT. The MLP used in learning average poses has 8 layers, and the hidden units are of the same size as the pose. LeakyReLU with a negative slope 0.2 is used as activation for all aforementioned MLPs.

\subsection{Hyperparameters} 

Our VQ-PAE is implemented in PyTorch~\cite{NEURIPS2019_9015}, and the experiments are performed on NVIDIA GeForce RTX 3090 GPU. We optimize the parameters of our network using the Adam optimizer~\cite{kingma2014adam}. We set the learning rate to $1\times 10^{-4}$ and the batch size to 32. We choose to use joint velocity in the local coordinate of the character as the input feature $\X$. We choose the size of the input motion sequence during training to correspond to 1 second. The exact size is different for different datasets depending on their framerate. For the hyperparameters used in frequency-scaled motion matching, we set $\lambda_1 = 0.5$ and $\lambda_2 = 1$.

\bibliographystyle{ACM-Reference-Format}
\bibliography{bibs}